\documentclass[10pt,journal,compsoc]{IEEEtran}
\ifCLASSOPTIONcompsoc
  
  \usepackage[nocompress]{cite}
\else
 
  \usepackage{cite}
\fi

\usepackage{times}
\usepackage{epsfig}
\usepackage{graphicx}
\usepackage{amsmath}
\usepackage{amssymb}
\usepackage{adjustbox,lipsum}
\usepackage[vlined,boxed,commentsnumbered]{algorithm2e}
\usepackage{subfigure}
\usepackage{algorithmic}
\usepackage{tablefootnote}

\ifCLASSINFOpdf

\usepackage{epsfig}
\usepackage{graphicx}
\usepackage{amsmath}
\usepackage{amssymb}
\usepackage{algorithmic}
\usepackage{adjustbox,lipsum}
\usepackage[vlined,boxed,commentsnumbered]{algorithm2e}
\usepackage{subfigure}
\usepackage{tablefootnote}
\usepackage{xcolor}
\usepackage[font=small,skip=5pt]{caption}
\else

\fi

\newtheorem{definition}{Definition}

\newcommand{\reals}[1]{\mathbb{R}^{#1}}
\newcommand{\enorm}[1]{\left\|{#1}\right\|}

\newcommand{\set}[1]{\left\{#1\right\}}

\newcommand{\half}{\frac{1}{2}}

\DeclareMathOperator*{\subjectto}{\text{subject to}}
\newcommand{\eye}[1]{\mathbf{I}_{#1}}

\renewcommand\cdots{...}

\newcommand{\suptensor}[1]{\mathfrak{S}^{d}}

\DeclareMathOperator*{\argmin}{arg\,min}

\newcommand{\comment}[1]{}

\newcommand{\done}[1]{}
\newcommand{\actodo}[1]{}

\newcommand{\seq}{X}

\newcommand*{\pseq}[1]{X^{\small{+}}_{{#1}}}
\newcommand*{\hpseq}[1]{\hat{X}^{\small{+}}_{{#1}}}
\newcommand*{\nseq}[1]{X^{\small{-}}_{{#1}}}
\newcommand*{\pdataset}{\mathcal{X}^{\small{+}}_{}}
\newcommand*{\ndataset}{\mathcal{X}^{\small{-}}_{}}
\newcommand*{\nfeat}[2]{\mathbf{x}^{{#1}\small{-}}_{{#2}}}
\newcommand*{\pfeat}[2]{\mathbf{x}^{{#1}\small{+}}_{{#2}}}

\newcommand*{\feat}{\mathbf{x}}
\newcommand*{\hfeat}{\mathbf{\hat{x}}}

\newcommand*{\ypseq}[1]{y^{\small{+}}_{{#1}}}

\newcommand{\card}[1]{\left|{#1}\right|}
\DeclareMathOperator*{\svmp}{SVMP}

\DeclareMathOperator*{\svm}{SVM}
\newcommand{\dataset}{\mathcal{X}}

\begin{document}
\title{Discriminative Video Representation Learning Using Support Vector Classifiers}

\author{Jue Wang \qquad
        Anoop Cherian \qquad
%         Fatih Porikli% <-this % stops a space
\IEEEcompsocitemizethanks{\IEEEcompsocthanksitem Jue Wang is with the Research School of Engineering, The
Australian National University, ACT 2601, Australia. E-mail: jue.wang@anu.edu.au\protect\\
\IEEEcompsocthanksitem Anoop Cherian is with Mistubishi Electric Research Labs (MERL), Cambridge, MA, E-mail: cherian@merl.com\protect\\
% \IEEEcompsocthanksitem Fatih Porikli is with The Australian National University, Canberra, Australia, E-mail:~fatih.porikli@anu.edu.au
}% <-this % stops a space
\thanks{}}

\markboth{TRANSACTIONS ON PATTERN ANALYSIS AND MACHINE INTELLIGENCE}%
{Shell \MakeLowercase{\textit{et al.}}: Discriminative Video Representation via Classifier Decision Boundary}

\IEEEtitleabstractindextext{%
\begin{abstract}
Most popular deep models for action recognition in videos generate independent predictions for short clips, which are then pooled heuristically to assign an action label to the full video segment. As not all frames may characterize the underlying action---indeed, many are common across multiple actions---pooling schemes that impose equal importance on all frames might be unfavorable. In an attempt to tackle this problem, we propose~\emph{discriminative pooling}, based on the notion that among the deep features generated on all short clips, there is at least one that characterizes the action. To identify these useful features, we resort to a negative bag consisting of features that are known to be irrelevant, for example, they are sampled either from datasets that are unrelated to our actions of interest or are CNN features produced via random noise as input. With the features from the video as a positive bag and the irrelevant features as the negative bag, we cast an objective to learn a (nonlinear) hyperplane that separates the unknown useful features from the rest in a multiple instance learning formulation within a support vector machine setup. We use the parameters of this separating hyperplane as a descriptor for the full video segment. Since these parameters are directly related to the support vectors in a max-margin framework, they can be treated as a weighted average pooling of the features from the bags, with zero weights given to non-support vectors. Our pooling scheme is end-to-end trainable within a deep learning framework. We report results from experiments on eight computer vision benchmark datasets spanning a variety of video-related tasks and demonstrate state-of-the-art performance across these tasks.
\end{abstract}

\begin{IEEEkeywords}
video representation, video data mining, discriminative pooling, action recognition, deep learning.
\end{IEEEkeywords}}

\maketitle
\IEEEdisplaynontitleabstractindextext

\IEEEpeerreviewmaketitle

\IEEEraisesectionheading{\section{Introduction}
\label{sec:intro}}
%Early trend used to represent the video by hand-crafted local features, such as HOG, HOF and %MBH~\cite{wang2011action,wang2013action}. However, t
\IEEEPARstart{W}{e} are witnessing an astronomical increase of video data around us. This data deluge has brought out the problem of effective video representation -- specifically, their semantic content -- to the forefront of computer vision research. The resurgence of convolutional neural networks (CNN) has enabled significant progress to be made on several problems in computer vision~\cite{he2016deep,he2017mask} and is now pushing forward the state-of-the-art in action recognition and video understanding as well~\cite{carreira2017quo,feichtenhofer2017spatiotemporal,Hu_2018_ECCV,zhou2017temporalrelation}. Even so, current solutions for video representation are still far from being practically useful, arguably due to the volumetric nature of this data modality and the complex nature of real-world human actions. %~\cite{carreira2017quo,feichtenhofer2017spatiotemporal,feichtenhofer2017temporal,feichtenhofer2016convolutional,hayat2015deep, simonyan2014two, simonyan2014very,Wang2016,,Si_2018_ECCV,Wang_2018_ECCV,Hu_2018_ECCV}.
\begin{figure}
	\begin{center}
        \includegraphics[width=1\linewidth,trim={0cm 0cm 0cm 0cm},clip]{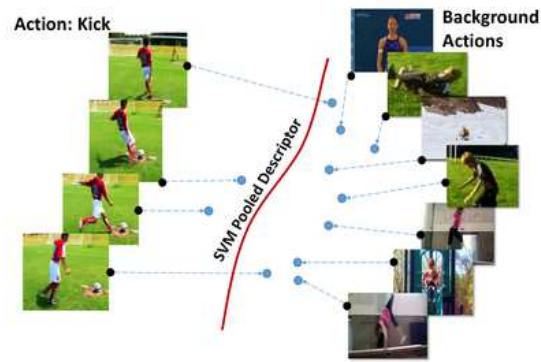}
	\end{center}
	\caption{A illustration of our discriminative pooling scheme. Our main idea is to learn a representation for the positive bag (left) of CNN features from the video of interest. To extract useful features from this video, we use a negative bag (right) of features from videos that are known to contain irrelevant/noise features. The representation learning problem is cast as a binary (non)-linear classification problem in an SVM setting; the hyperplane found via the optimization (which is a linear combination of support vectors) is used as the representation of the positive bag, which we call the~\emph{SVM pooled descriptor}.}
	\label{fig:1}
\end{figure}

% \begin{figure}
% 	\begin{center}
% 		%\includegraphics[width=1\linewidth,trim={3cm 0cm 1cm 0cm},clip]{figure/figure1.jpg}
%         \includegraphics[width=1\linewidth,trim={0cm 0cm 0cm 0cm},clip]{figure/Figures4.jpg}
% 	\end{center}
% 	\caption{A visualization of discriminative pooling applied to RGB frames in a sequence. (i) a sample frame, (ii) average pooling all frames, (iii) dynamic image by rank pooling~\cite{bilen2016dynamic}, and (iv) our SVM pooling. Our representation captures more details of the actions as it learns to discriminate parts of the foreground against a background set. In this case, we used the average-pooled frames as the background. }
% 	\label{fig:rgb}
% \end{figure}
%, and a classifier may be confused by predictions from background actions

Using effective architectures, CNNs are often found to extract features from images that perform well on recognition tasks. Leveraging this know-how, deep learning solutions for video action recognition have so far been straightforward extensions of image-based models\cite{simonyan2014two,ji20133d,zhou2017temporalrelation}. However, applying such models directly on video data is not an easy task as the video can be arbitrarily long, to address which a CNN may need to be scaled up by yet another dimension of complexity, which could increase the number of parameters sharply. This demands more advanced computational infrastructures and greater quantities of clean training data~\cite{carreira2017quo,monfort2018moments}. To overcome this problem, the trend has been on converting the video data to short temporal segments consisting of one to a few frames, on which the existing image-based CNN models are trained. For example, in the popular two-stream model~\cite{feichtenhofer2016convolutional, simonyan2014two, simonyan2014very, wang2015action, wangtwo}, the CNNs are trained to independently predict actions from short video clips (consisting of single frames or stacks of about ten optical flow frames) or a snippet of about 64 frames as in the recent I3D architecture~\cite{carreira2017quo}; these predictions are then pooled to generate a prediction for the full sequence -- typically using average/max pooling. While average pooling gives equal weights to all the predictions, max pooling may be sensitive to outliers. There have also been recent approaches that learn representations over features produced by, say a two-stream model, such as the temporal relation networks of Zhou et al.~\cite{zhou2017temporalrelation}, the rank pooling and its variants Bilen et al.,~\cite{bilen2016dynamic}, Fernando et al.,~\cite{fernando2015modeling}, and Cherian et al.,~\cite{cherian2018non,grp} that capture the action dynamics, higher-order statistics of CNN features Cherian et al.,~\cite{cherian2017second,cherian2017higher}, CNN features along motion trajectories Wang et al., ~\cite{wang2015action} and temporal segments Wang et al.,~\cite{Wang2016}, to name a few. However, none of these methods  avoid learning meaningless information from the noise/background within the video, explicitly modeling which and demonstrating its benefits, are the main contributions of this paper.
% and Carreira et al.~\cite{carreira2017quo} achieve state-of-the-art performance by using inflated 3D kernel and pre-training the network with large-scale datasets. part from that, several works also try to tackle this problem by using different pooling strategies: for example,

%To this end, we observe that capturing long-term dynamics would be essential for video representation and not all predictions on the short video snippets are equally informative, yet some of them must be. 
To this end, we observe that not all predictions on the
short video snippets are equally informative, yet some of
them must be~\cite{schindler2008action}. This allows us to cast the problem in a multiple instance learning (MIL) framework, where we assume that some of the features in s given sequence are indeed useful, while the rest are not. We assume all the CNN features from a sequence (containing both the good and the bad features) to represent a positive bag, while CNN features from unrelated video frames or synthetically generated random noise frames as a negative bag. We would ideally want the features in the negative bag to be correlated well to the uninformative features in the positive bag. We then formulate a binary classification problem of separating as many good features as possible in the positive bag using a discriminative classifier (we use a support vector machine (SVM) for this purpose). The decision boundary of this classifier thus learned is then used as a descriptor for the entire video sequence, which we call the SVM Pooled (SVMP) descriptor. To accommodate the fact that we are dealing with temporally-ordered data in the positive bag, we also explore learning our representations with partial ordering relations. An illustration of our SVMP scheme is shown in the Figure~\ref{fig:1}.

% This idea has a wide range of capacity for the input data. It could be any kind of video information such as RGB frame, optical flow, skeleton coordinate or high level CNN feature map from any intermediate layer. The core idea is to characterize the discriminative information of the video and summarize it into one descriptor. Additionally, our SVMP scheme is also found to work well in the image sets problem, such as image set verification, where we learn a discriminative classifier for a bag of images. Of course, in this case, the temporal order is not necessarily added.

Our SVMP scheme/descriptor shares several properties of standard pooled descriptors, however also showcases several important advantages. For example, similar to other pooling schemes, SVM pooling results in a compact and fixed length representation of videos of arbitrary length. However differently, our pooling gives different weights to different features, and thus may be seen as a type of weighted average pooling, by filtering out features that are perhaps irrelevant for action recognition. Further, given that our setup uses a max-margin encoding of the features, the pooled descriptor is relatively stable with respect to data perturbations and outliers. Our scheme is agnostic to the feature extractor part of the system, for example, it could be applied to the intermediate features from any CNN model or even hand-crafted features. Moreover, the temporal dynamics of actions are explicitly encoded in the formulation. The scheme is fast to implement using publicly available SVM solvers, and also could be trained in an end-to-end manner within a CNN setup.

% It is robust to classifier outliers thanks to the SVM formulation. 4. It can be applied on any type of video data while keeping the original size, for example, when applying on RGB frames, the SVMP descriptor will still be an image. 5. It is computationally cheap, compared with other popular pooling method. 6. The SVMP descriptor is compatible for different classifier including kernelized SVM or multi-layer perceptron (MLP). 
% To provide intuitions, in Figure~\ref{fig:rgb}, we provide a visualization of our descriptor applied directly on video frames while also applying other pooling schemes~\cite{bilen2016dynamic} as comparisons. As is clear, SVMP captures the essence of action dynamics in more detail in comparison to prior works.

To evaluate our SVMP scheme, we provide extensive experiments on various datasets spanning a diverse set of tasks, namely action recognition and forecasting on HMDB-51~\cite{kuehne2011hmdb}, UCF-101~\cite{soomro2012ucf101}, Kinetics-600~\cite{kay2017kinetics} and Charades~\cite{sigurdsson2016hollywood}; skeleton-based action recognition on MSR action-3D~\cite{li2010action}, and NTU-RGBD~\cite{shahroudy2016ntu}; image-set verification on the PubFig dataset~\cite{kumar2009attribute}, and video-texture recognition on the YUP++ dataset~\cite{feichtenhofer2017temporal}. We outperform standard pooling methods on these datasets by a significant margin (between 3--14\%) and demonstrate superior performance against state-of-the-art results by 1--5\%.

Before moving on, we summarize below the main contributions of this paper:
\begin{itemize}
\item We introduce the concept of multiple instance learning (MIL) into a binary SVM classification problem for learning video descriptors.
\item We propose SVM pooling that captures and summarizes the discriminative features in a video sequence while explicitly encoding the action dynamics.
\item We explore variants of our optimization problem and present progressively cheaper inference schemes, including a joint pooling and classification objective, as well as an end-to-end learnable CNN architecture.
\item We demonstrate the usefulness of our SVMP descriptor by applying it on eight popular vision benchmarks spanning diverse input data modalities and CNN architectures.
\end{itemize}

\section{Related Work}
\label{sec:related_work}
% Prior works for representing sets of images (including videos) essentially consists of two steps, namely (i) to either represent 
% The research topic for solving group-image based classification problems mainly include action recognition in videos, dynamic scene recognition and image sets matching, which contain many group of images and frames in one group share the same label. The standard way to solve this problem involves two major steps: 1. to find a representation for the group or for each frame in the group. 2. to define suitable  distance metrics to compute the similarity between these representations. Note that: a late fusion strategy is required if the representation is frame level. Based on the type of the presentation, existing methods can be roughly divided into two classes: deep method and non-deep method.
The problem of video representation learning has received significant interest over the past decades. Thus, we restrict our literature review to some of the more recent methods, and defer the interested reader to excellent surveys on the topics such as~\cite{herath2017going,poppe2010survey,aggarwal2011human}.
\subsection{Video Representation Using Shallow Features}
Traditional methods for video action recognition typically use hand-crafted local features, such as dense trajectories, HOG, HOF, etc.~\cite{wang2011action}, which model videos by combining dense sampling with feature tracking. However, the camera motion, as one of the video natures, usually result in non-static video background and hurt the quality of features. To tackle this problem, Wang et al.~\cite{wang2013action} improved the performance of dense trajectories by removing background trajectories and warping optical flow. Based on the improved dense trajectories, high-level representations are designed via pooling appearance and flow features along these trajectories, and have been found to be useful to capture human actions. For example, Sadanand et al.~\cite{sadanand2012action} propose Action Bank, which converts the individual action detector into semantic and viewpoint space. Similarly, Bag of words model~\cite{sivic2003video}, Fisher vector~\cite{perronnin2010improving}, and VLAD~\cite{jegou2012aggregating} representations are mid-level representations built on such hand-crafted features with the aim of summarizing local descriptors into a single vector representation. In Peng et al.~\cite{peng2016bag}, a detailed survey of these ideas is presented. In comparison to these classic representation learning schemes, our proposed setup is grounded on discriminatively separating useful data from the rest.

\subsection{Video Representation Using Deep Features}With the resurgence of deep learning methods for object recognition~\cite{krizhevsky2012imagenet}, there have been several attempts to adapt these models to action recognition. Recent practice is to feed the video data, including RGB frames, optical flow subsequences, and 3D skeleton data into a deep (recurrent) network to train it in a supervised manner. Successful methods following this approach are the two-stream models and their extensions~\cite{feichtenhofer2017spatiotemporal,feichtenhofer2017temporal,hayat2015deep,kim2017interpretable,simonyan2014two}. As apparent from its name, it has two streams, spatial stream is to capture the appearance information from RGB frames and temporal stream is to learn the motion dynamics from stacked optical flow. And then, they apply early or late fusion strategy to predict the final label. Although the architecture of these networks are different, the core idea is to split the video into short clips and embed them into a semantic feature space, and then recognize the actions either by aggregating the individual features per clip using some statistic (such as max or average) or directly training a CNN based end-to-end classifier~\cite{feichtenhofer2017spatiotemporal}. While the latter schemes are appealing, they usually need to store the feature maps from all the frames in memory which may be prohibitive for longer sequences. Moreover, this kind of training strategy may fail to capture the long-term dynamics in the video sequence. To tackle this problem, some recurrent models~\cite{baccouche2011sequential,donahue2015long,du2015hierarchical,li2016action,srivastava2015unsupervised,yue2015beyond} are proposed, which use long-short term memory (LSTM) or gate recurrent unit (GRU) to embed the temporal relationship among frames by using logistic gates and hidden states. However, the recurrent neural networks are usually hard to train~\cite{pascanu2013difficulty} due to the exploding and vanishing gradient problem. Temporal Segment Network (TSN)~\cite{Wang2016} and Temporal Relation Network (TRN)~\cite{zhou2017temporalrelation} provide alternative solutions that are easier to train.
% This is because recurrent models are trained by back-propagation through time, and therefore unfolded into feed forward networks with multiple layers. When gradient is passed back through many time steps, it tends to grow or vanish. Even with good initial weights, it is still seen as a difficult task to effectively train such a system towards state-of-the-art performance.

Another promising solution is to use 3D convolutional filters~\cite{carreira2017quo,tran2015learning,wang2018non,zhou2018mict,wang2017appearance}. Compared to 2D filters, 3D filters can capture both spatial and temporal video structure. However, feeding the entire video into the CNNs may be computationally prohibitive. Further, 3D kernels bring more parameters into the architecture; as a result, may demand large and clean data for effective training~\cite{carreira2017quo}. While, an effective CNN architecture that can extract useful action-related features is essential to make progress in video understanding, we focus on the other aspect of the problem -- that is, given a CNN architecture how well can we summarize the features it produces for improving action recognition. To this end, our efforts in this paper can be seen as complimentary to these recent approaches.   %In contrast to these approaches, we cast this problem into a pooling scheme of mapping the correct set of frames automatically into a hyperplane that is discriminative for video representation. A scheme that has the similar motivation of ours is the recent work of Wang et al.,~\cite{Wang2016}, however they use manually-defined video segmentation for equally-spaced snippet sampling. 

%Besides, it seems appealing by using more source of input in multiple streams. More recent works try to find cue from Depth feature, RGB difference, and multi-view images, which may require more effort for collection and pre-processing.~\cite{Garcia_2018_ECCV,wang2018temporal,Hu_2018_ECCV,wwang_2018_ECCV}

% Note that such a hyperplane is of the same dimensionality as the data and well-known as a weighted combination of each data point, where the weight captures how discriminative each point is.

% To this end, some pooling schemes~\cite{grp,fernando2015modeling,dynamic_flow,bilen2016dynamic,cherian2018non} are proposed based on this design. This paper is also motivated by the previous pooling schemes. 

\subsection{Video Representation Using Pooling Schemes}Typically, pooling schemes consolidate input data into compact representations based on some data statistic that summarizes the useful content. For example, average and max pooling captures zero-th and first order statistics. There are also works that use higher-order pooling, such as Cherian and Gould~\cite{anoop_secondorder} using second-order, Cherian et al.~\cite{cherian2017higher} using third-order, and Girdhar et al.,~\cite{Girdhar_17a_ActionVLAD} proposing a video variant of the VLAD encoding which is approximately a mixture model. A recent trend in pooling schemes, which we also follow in this paper, is to use the parameters of a data modeling function, as the representation. For example, rank pooling~\cite{fernando2015modeling} proposes to use the parameters of a support vector regressor as a video representation. In Bilen et al.,~\cite{bilen2016dynamic}, rank pooling is extended towards an early frame-level fusion, dubbed~\emph{dynamic images}; Wang et al.~\cite{dynamic_flow}, extends this idea to use optical flow, which they call \emph{dynamic flow} representation. Cherian et al.~\cite{grp} generalized rank pooling to include multiple hyperplanes as a subspace, enabling a richer characterization of the spatio-temporal details of the video. This idea was further extended to non-linear feature representations via kernelized rank pooling in \cite{cherian2018non}. However, while most of these methods optimize a rank-SVM based regression formulation, our motivation and formulation are different. We use the parameters of a binary SVM to be the video level descriptor, which is trained to classify the frame level features from a pre-selected (but arbitrary) bag of negative features. Similar works are Exemplar-SVMs~\cite{malisiewicz2011ensemble,willems2009exemplar,zepeda2015exemplar}, that learn feature filters per data sample and then use these filters for feature extraction. However, in this paper, we use the decision boundary of the SVM to be the video level descriptor, that separate as many discriminative features as possible in each sequence while implicitly encoding the temporal order of these features.

\begin{figure*}[ht]
	\begin{center}
        \includegraphics[width=0.7\linewidth,trim={0cm 0cm 0cm 0cm},clip]{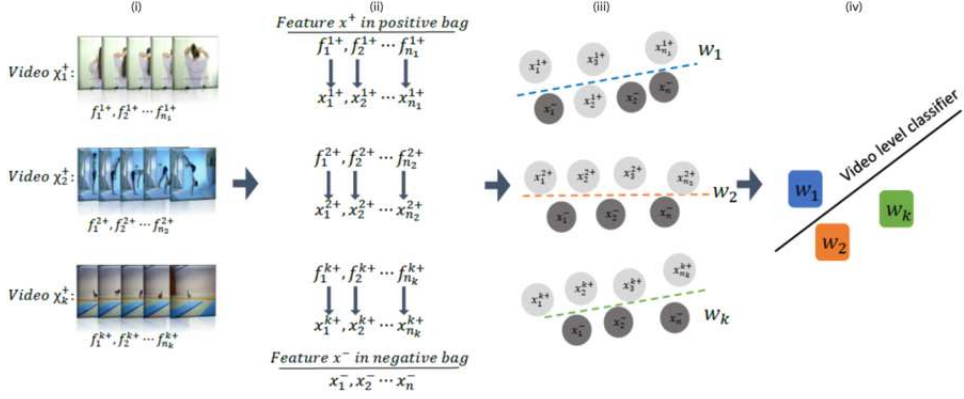}
	\end{center}
	\caption{Illustration of our SVM Pooling scheme. (i) Extraction of frames from videos, (ii) Converting frames $f$ into feature $x$, (iii) Learning decision boundary $w$ from feature $x$, and (iv) Using $w$ as video descriptor.} %We take the frame level feature from both positive and negative bags to learn a decision boundary as the video representation.}
   	\label{fig:2}
\end{figure*}

\subsection{Multiple Instance Learning} An important component of our algorithm is the MIL scheme, which is a popular data selection technique~\cite{cinbis2017weakly,li2015multiple,wu2015deep,yi2016human,zhang2015self}. In the context of video representation, schemes similar in motivation have been suggested before. For example, Satkin and Hebert~\cite{satkin2010modeling} explore the effect of temporal cropping of videos to regions of actions; however, it assumes these regions are continuous. Nowozin et al.~\cite{nowozin2007discriminative} represent videos as sequences of discretized spatiotemporal sets and reduces the recognition task into a max-gain sequence finding problem on these sets using an LPBoost classifier. Similar to ours, Li et al.~\cite{li2013dynamic} propose an MIL setup for complex activity recognition using a dynamic pooling operator -- a binary vector that selects input frames to be part of an action, which is learned by reducing the MIL problem to a set of linear programs. Chen and Nevatia~\cite{sun2014discover} propose a latent variable based model to explicitly localize discriminative video segments where events take place. Vahdat et al. present a compositional model in~\cite{vahdat2013compositional} for video event detection, which is presented using a multiple kernel learning based latent SVM. While all these schemes share similar motivations as ours, we cast our MIL problem in the setting of normalized set kernels~\cite{gartner2002multi} and reduce the formulation to standard SVM setup which can be solved rapidly. In the $\propto$-SVMs of Yu et al.,~\cite{lai2014video,yu2013propto}, the positive bags are assumed to have a fixed fraction of positives, which is a criterion we also assume in our framework. However, the negative bag selection, optimization setup and our goals are different; specifically, our goal is to learn a video representation for any subsequent task including recognition, anticipation, and detection, while the framework in~\cite{lai2014video} is designed for event detection. And we generate the negative bag by using CNN features generated via inputing random noise images to the network.

The current paper is an extension of our published conference paper~\cite{SVMP} and differs in the following ways. Apart from the more elaborate literature survey we present, we also provide extensions of our pooling scheme, specifically by incorporating temporal-ordering constraints. We provide detailed derivations of our end-to-end pooling variant. We further present elaborate experiments on five more datasets in addition to the three datasets that we used in~\cite{SVMP}, including a large scale action recognition experiment using the recently proposed Kinetics-600 dataset. %We also present extensive ablative studies furnishing deeper insights into the proposed scheme.
% \begin{itemize}
% \item We provide more extensive overview for the related work in terms of the video representation.
% \item We make more analysis and provide new optimization solution.
% \item We put new temporal ranking constraint into the formulation for better video representation.
% \item We provide more details in the end-to-end learning set-up as well as an end-to-end learnable attention scheme to simulate the SVMP.
% \item We evaluate the performance over five more dataset using both deeper CNN architecture and non-deep methods as the feature encoder.
% \item We extend our SVMP scheme from video level representation to image set representation, demonstrating a wide range of applications of our algorithm.
% \item We define an new video representation, namely SVMP image, which shows better performance compared to previous dynamic image~\cite{bilen2016dynamic}.
% \item We give a detailed discussion to give more insight about why this scheme could work well in different tasks.
% \end{itemize}

\section{Proposed Method}
%\label{sec:proposed_method}
\label{sec:setup}
In this section, we first describe the problem of learning SVMP descriptors and introduce three different ways to solve it. Before proceeding, we provide a snapshot of our main idea and problem setup graphically in Figure~\ref{fig:2}. Starting from frames (or flow images) in positive and negative bags, these frames are first passed through some CNN model for feature generation. These features are then passed to our SVMP module that learns (non-linear) hyperplanes separating the features from the positive bag against the ones from the negative bag, the latter is assumed fixed for all videos. These hyperplane representations are then used to train an action classifier at the video level. In the following, we formalize these ideas concretely.
% First, we transfer both positive bag and negative bag into the feature space. Note that the feature here has a wide range of selection including deep CNN features or local hand-crafted features. And then, we apply our SVMP scheme to learn hyperplanes for each sequence. These hyperplanes will be treated as the video descriptor to do further tasks, such as the video classification in this Figure. In the following, we will formulate the problem properly. 

\subsection{Problem Setup}
\label{sec:setup}
Let us assume we are given a dataset of $N$ video sequences $\pdataset = \set{\pseq{1}, \pseq{2},\cdots, \pseq{N}}$, where each $\pseq{i}$ is a set of frame level features, $i.e.$, $\pseq{i}=\set{\pfeat{i}{1}, \pfeat{i}{2}, \cdots, \pfeat{i}{n}}$, each $\pfeat{i}{k}\in\reals{p}$. We assume that each $\pseq{i}$ is associated with an action class label $\ypseq{i}\in\set{1,2,\cdots, d}$. Further, the $+$ sign denotes that the features and the sequences represent a positive bag. We also assume that we have access to a set of sequences $\ndataset=\set{\nseq{1}, \nseq{2},\cdots \nseq{M}}$ belonging to actions different from those in $\pdataset$, where each $\nseq{j}=\set{\nfeat{j}{1}, \nfeat{j}{2}, \cdots, \nfeat{j}{n}}$ are the features associated with a negative bag, each $\nfeat{j}{k}\in\reals{p}$. For simplicity, we assume all sequences have same number $n$ of features. Further note that our scheme is agnostic to the type of features, i.e., the feature may be from a CNN or are hand-crafted. 
% Let us assume we are given a dataset of $N$ video sequences $\pdataset = \set{\pseq{1}, \pseq{2},\cdots, \pseq{N}}$, where each $\pseq{i}$ is a set of  frame level features $\set{\pfeat{i}{1}, \pfeat{i}{2}, \cdots, \pfeat{i}{n}}$, each $\pfeat{i}{k}\in\reals{p}$ and $\feat_{k}=\mathcal{T}(\fram_{k})$, where $\mathcal{T}(\cdot) $ is the frame representation function (which could be a CNN feature extractor or one that generates hand-crafted features). We assume that each $\pseq{i}$ is associated with a class label $\ypseq{i}\in\set{1,2,\cdots, d}$. Further, the $+$ sign denotes that the features and the sequences represent a positive bag. We also assume that we have access to a set of sequences $\ndataset=\set{\nseq{1}, \nseq{2},\cdots \nseq{M}}$ belonging to classes different from those in $\pdataset$, where each $\nseq{j}=\set{\nfeat{j}{1}, \nfeat{j}{2}, \cdots, \nfeat{j}{n}}$ are the features associated with a negative bag, each $\nfeat{j}{k}\in\reals{p}$. For simplicity, we assume all sequences have same number $n$ of features.

Our goals are two-fold, namely (i) to learn a classifier decision boundary for every sequence in $\pdataset$ that separates a fraction $\eta$ of them from the features in $\ndataset$ and (ii) to learn video level classifiers on the classes in the positive bags that are represented by the learned decision boundaries in (i). In the following,  we will provide a multiple instance learning formulation for achieving (i), and a joint objective combining (i) and learning (ii). However, before presenting our scheme, we believe it may be useful to gain some insights into the main motivations for our scheme. 

\begin{figure}[htbp]
\subfigure[]{\label{fig:3a}\includegraphics[width=4cm,trim={4cm 2cm 15cm 6cm},clip]{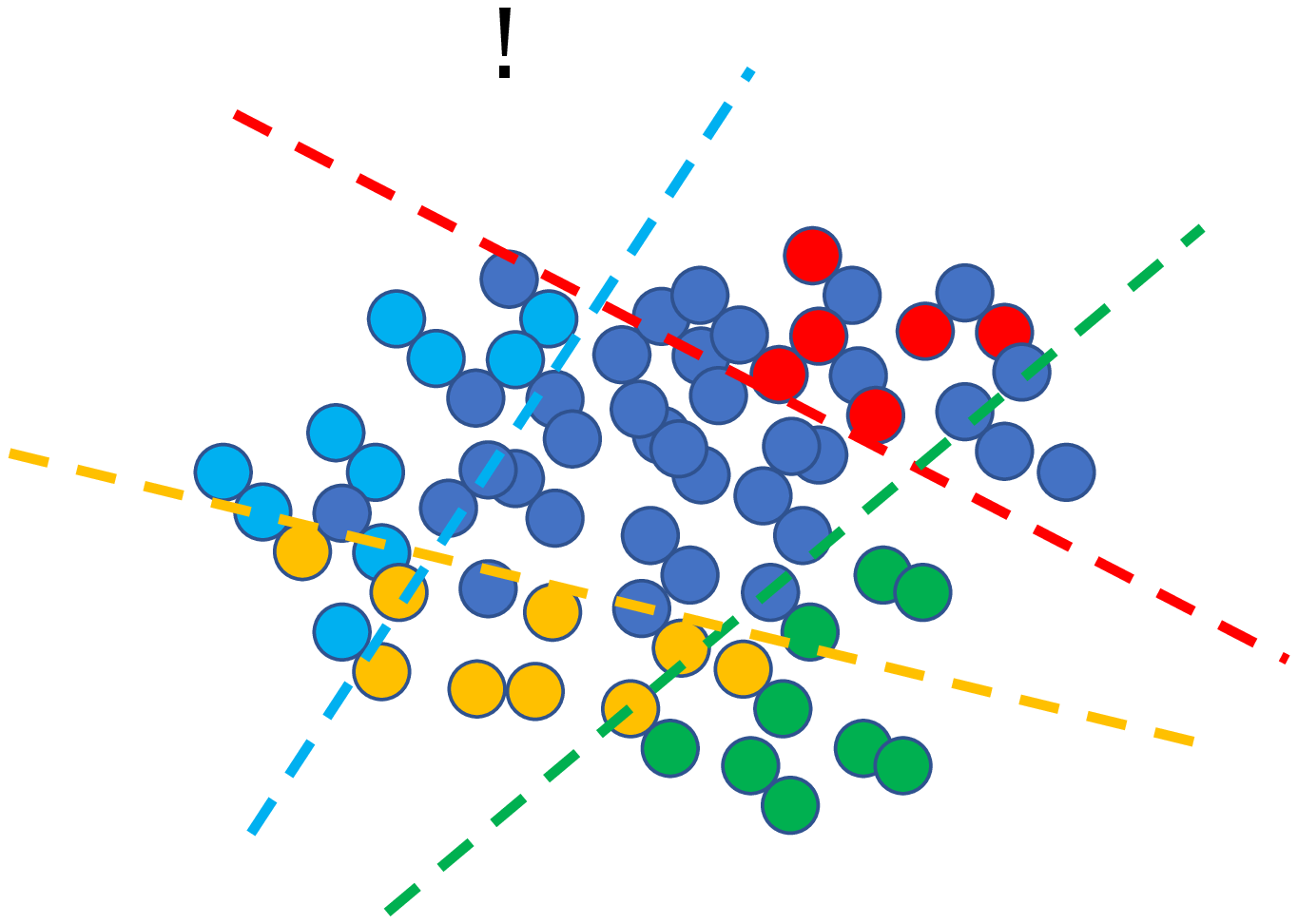}}
\subfigure[]{\label{fig:3b}\includegraphics[width=4cm,trim={4cm 2cm 15cm 6cm},clip]{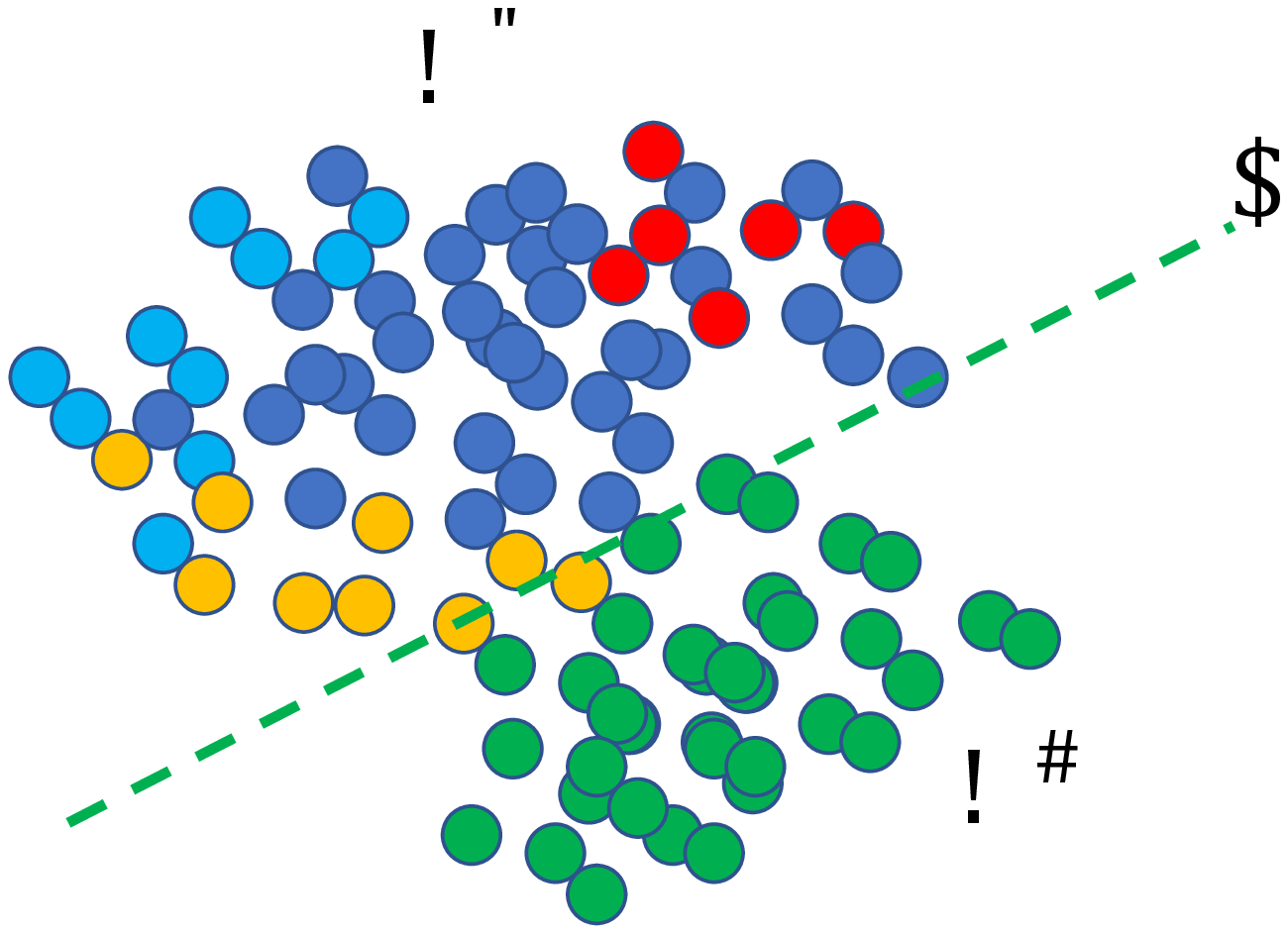}}
\caption{An illustration of our overall idea. (a) the input data points, and the plausible hyperplanes satisfying some $\eta$ constraint, (b) when noise $\mathcal{X}^{-}$ is introduced (green dots), it helps identify noisy features/data dimensions, towards producing a hyperplane $w$ that classifies useful data from noise, while satisfying the $\eta$ constraint. }
\label{fig:3}
\end{figure}

As alluded to above, given the positive and negative bags, our goal is to learn a linear (or non-linear) classification boundary that could separate the two bags with a classification accuracy of $\eta\%$ -- this classification boundary is used as the descriptor for the positive bag. Referring to the conceptual illustration in Figure~\ref{fig:3a}, when no negative bag is present, there are several ways to find a decision hyperplane in a max-margin setup that could potentially satisfy the $\eta$ constraint. However, there is no guarantee that these hyperplanes are useful for action recognition. Instead, by introducing a negative bag, which is almost certainly to contain irrelevant features, it may be easier for the decision boundary to identify useless features from the rest; the latter containing useful action related features, as shown in Figure~\ref{fig:3b}. This is precisely our intuitions for proposing this scheme. 

% we first introduce the frame representation function $\mathcal{T}(\cdot) $. After that,
% \subsection{Independent Frame Representation}
% As is shown in the Figure~\ref{fig:2}, we first extract frames $\set{\pfram{i}{1}, \pfram{i}{2}, \cdots, \pfram{i}{n}}$ from sequences in the positive bag. Then we generate the frame level feature $\set{\pfeat{i}{1}, \pfeat{i}{2}, \cdots, \pfeat{i}{n}}$ for each of them. Theoretically, the transfer function $\mathcal{T}(\cdot)$ could be anything. For better comparison and performance, we apply the state of the art feature learning/representation methods including CNN features, local hand-crafted features and Lie group representation for various original data source such as RGB frame, optical flow and skeleton data. The feature $\pfeat{i}{k}$ will become the input to our SVM pooling scheme.

\subsection{Learning Decision Boundaries}

As described above, our goal in this section is to generate a descriptor for each sequence $\pseq{}\in\pdataset$; this descriptor we define to be the learned parameters of a hyperplane that separates the features $\pfeat{}{}\in\pseq{}$ from all features in $\ndataset$. We do not want to warrant that all $\pfeat{}{}$ can be separated from $\ndataset$ (since several of them may belong to a background class), however we assume that at least a fixed fraction $\eta$ of them are classifiable. Mathematically, suppose the tuple $(w_i,b_i)$ represents the parameters of a max-margin hyperplane separating some of the features in a positive bag $\pseq{i}$ from all features in $\ndataset$, then we cast the following objective, which is a variant of the sparse MIL (SMIL)~\cite{bunescu2007multiple}, normalized set kernel (NSK)~\cite{gartner2002multi}, and $\propto$-SVM~\cite{yu2013propto} formulations:
\begin{align}
\label{eq:mil}
&\!\!\argmin_{w_i\in\reals{p},b_i\in\reals{},\zeta\geq 0} P1(w_i,b_i) := \half\enorm{w_i}^2 +  C_1\sum_{k=1}^{(M+1)n}\zeta_k\\
&\subjectto\ \theta(\feat;\eta)\left(w_i^T\feat+b_i\right)  \geq 1 - \zeta_k\\
\label{eq:5}&\theta(\feat;\eta)= -1, \forall \feat \in \left\{\pseq{i}\bigcup \ndataset\right\}\backslash \hpseq{i}\\
\label{eq:6}&\theta(\hfeat; \eta) = 1, \forall \hfeat \in\hpseq{i}  \\
&\card{\hpseq{i}}  \geq \eta \card{\pseq{i}}. \quad\quad
\label{eq:ratio-constraint} 
\end{align} 
%\sum_{\feat\in\pseq{i}\bigcup \ndataset} 
In the above formulation, we assume that there is a subset $\hpseq{i}\subset\pseq{i}$ that is classifiable, while the rest of the positive bag need not be, as captured by the ratio in~\eqref{eq:ratio-constraint}. The variables $\zeta$ capture the non-negative slacks weighted by a regularization parameter $C_1$, and the function $\theta$ provides the label of the respective features. Unlike SMIL or NSK objectives, that assumes the individual features $\feat$ are summable, our problem is non-convex due to the unknown set $\hpseq{}$. However, this is not a serious deterrent to the usefulness of our formulation and can be tackled as described in the sequel and supported by our experimental results.

As our formulation is built on an SVM objective, we call this specific discriminative pooling scheme as~\emph{SVM pooling} and formally define the descriptor for a sequence as:
\begin{definition}[SVM Pooling Desc.]
\label{def:svmp}
Given a sequence $\seq$ of features $\feat\in\reals{p}$ and a negative dataset $\ndataset$, we define the~\emph{SVM Pooling} (SVMP) descriptor as $\svmp(\seq) = [w,b]^T\in\reals{p+1}$, where the tuple $(w,b)$ is obtained as the solution of problem $P1$ defined in~\eqref{eq:mil}.
\end{definition}

\subsection{Optimization Solutions}
\label{opti_solution}
The problem $P1$ could be posed as a mixed-integer quadratic program (MIQP), which is unfortunately known to be in NP~\cite{lazimy1982mixed}. The problem $P1$ is also non-convex due to the proportionality constraint $\eta$, and given that the labels $\theta(\feat;\eta)$ are unknown. Towards a practically useful approximate solution circumventing these difficulties, we present three optimization strategies below.
\subsubsection{Exhaustive Enumeration}
A na\"ive way to solve problem $P1$ could be to enumerate all the possible $\theta(\feat;\eta)$ that meet a given $\eta$ constraint, which reduces solving the problem $P1$ to the classical SVM problem for each instantiation of the plausible $\theta$ assignments. In such a setting, for a given sequence, we can rewrite~\eqref{eq:mil} as:
\begin{align}
\argmin_{w_i\in\reals{p},b_i\in\reals{},\zeta\geq 0} &\half\enorm{w_i}^2 +  C_1\!\!\!\sum_{k=1}^{(M+1)n}\zeta_k\notag \\  &+ \max(0,1-\zeta_k-\theta(\feat;\eta)(w_i^T\feat+b_i)),
\end{align} 
where the constraints are included via the hinge loss. Once these subproblems are solved, we could compare the optimal solutions for the various subsets of the positive bag, and pick the best solution with smallest objective value. As is apparent, this na\"ive strategy becomes problematic for longer sequences or when $\eta$ is not suitably chosen.  
% Alg.~\ref{alg1} describes this strategy.
% \begin{algorithm}  
% 	\SetAlgoLined
% 	\KwIn{$\pseq{}$, $\ndataset$, $\eta$}
% 	$Obj_{min} = \infty,\ w_{min}= \infty,\ b_{min}=\infty$;\\
% 	\ForAll{$\theta$ satisfying $\card{\hpseq{}}  \geq \eta \card{\pseq{}}$}{
%     	$Obj,\ w,\ b \leftarrow \min_{w,b} \svm(\hpseq{},\ \ndataset,\ \theta)$\;
%         \ \ \If{$Obj\leq Obj_{min}$}{$w_{min}=w,\ b_{min}=b$\;}        
%         } 
% 	\KwRet{$[w_{min},b_{min}]$}
% 	\caption{Enumerative solution to the MIL problem $P1$}
% 	\label{alg1}
% \end{algorithm}

\subsubsection{Alternating algorithm}
\label{alternating}
This is a variant of the scheme proposed in~\cite{yu2013propto}. Instead of enumerating all possible $\theta(\feat;\eta)$ as above, the main idea here is to fix $\theta(\feat;\eta)$ or $[w,b]$ alternately and optimize the other. The detailed algorithm is shown in the Alg.~\ref{alg2}. 

\begin{algorithm}  
	\SetAlgoLined
	\KwIn{$\pseq{}$, $\ndataset$, $\eta$}
	$Initialize\ \theta\ according\ to\ \eta$\;
	\Repeat{$Reduction\ is\ smaller\ than\ a\ threshold\ (10^{-4})$}{$Fix\ \theta\ to\ solve\ [w,b]\leftarrow\svm(\pseq{},\ \ndataset,\ \theta)$\;
    $Fix\ [w,b]\ to\ solve\ \theta\colon$
    $\ \ Reinitialize\ \theta_i \leftarrow -1,\forall i\in(i,n)$\;
    \ \ \For{$i=1\ \to\ n$}{$Set\ \theta_i \leftarrow 1;$\\ $record\ the\ reduction\ of\ Objective$}
    $\ \ Sort\ and\ select\ the\ top\ R\ reductions,\ R=\eta n$\;
    $\ \ Get\ \theta\ according\ to\ the\ sorting$;}
	\KwRet{$[w,b]$}
	\caption{Alternating solution to the MIL problem $P1$}
	\label{alg2}
\end{algorithm}

In the Algorithm~\ref{alg2}, fixing $\theta$ to solve $[w,b]$ is a standard SVMP problem as in the enumeration algorithm above. When fixing $[w,b]$ to solve $\theta$, we apply a similar strategy as in~\cite{yu2013propto}; i.e., to initialize all labels in $\theta$ as $-1$, and then to turn each $\theta_i$ to $+1$ and record the reduction in the objective. Next, we sort these reductions to get the top $R$ best reductions, where $R=\eta n$. A higher reduction implies it may lead to a smaller objective. Next, these top $R$ $\theta_i$ will be set to $+1$ in $\theta$. While, there is no theoretical guarantee for this scheme to converge to a fixed point, empirically we observe a useful convergence, which we limit via a suitable threshold.%From theoretical perspective, there is no guarantee that this would converge. So, we stop the optimization until the reduction is no longer decreasing (or decrease within a small margin).
% Apparently, both algorithms are not computationally cheap, and could take $\mathcal{O}(n^{\eta n})$ times in enumeration algorithm and $\mathcal{O}(n\log{}n)$ in each iteration of alternating algorithm. 
\subsubsection{Parameter-tuning algorithm}
As is clear, both the above schemes may be computationally expensive in general. We note that the regularization parameter $C_1$ in~\eqref{eq:mil} controls the positiveness of the slack variables $\zeta$, thereby influencing the training error rate. A smaller value of $C_1$ allows more data points to be misclassified. If we make the assumption that useful features from the sequences are easily classifiable compared to background features, then a smaller value of $C_1$ could help find the decision hyperplane easily (further assuming the negative bag is suitably chosen). However, the correct value of $C_1$ depends on each sequence. Thus, in Algorithm~\eqref{alg3}, we propose a heuristic scheme to find the SVMP descriptor for a given sequence $\pseq{}$ by iteratively tuning $C_1$ such that at least a fraction $\eta$ of the features in the positive bag are classified as positive. 

\begin{algorithm}  
	\SetAlgoLined
	\KwIn{$\pseq{}$, $\ndataset$, $\eta$}
	$C_1 \leftarrow \epsilon,\ \lambda > 1$\;
	\Repeat{$\frac{\card{\hpseq{}}}{\card{\pseq{}}}\geq \eta$} {
		$C_1 \leftarrow \lambda C_1$\;
		$[w,b] \leftarrow \argmin_{w,b} \svm(\pseq{},\ \ndataset,\ C_1)$\;
		$\hpseq{} \leftarrow \set{\feat\in\pseq{}\ |\ w^T\feat+b \geq 0}$\;
	}
	\KwRet{$[w,b]$}
	\caption{Parameter-tuning solution for MIL problem $P1$}
	\label{alg3}
\end{algorithm}

\emph{A natural question here is how optimal is this heuristic?} Note that, each step of Algorithm~\eqref{alg3} solves a standard SVM objective. Suppose we have an oracle that could give us a fixed value $C$ for $C_1$ that works for all action sequences for a fixed $\eta$. As is clear, there could be multiple combinations of data points in $\hpseq{}$ that could satisfy this $\eta$ (as we explored in the Enumeration algorithm above). If $\hpseq{p}$ is one such $\hpseq{}$. Then, $P1$ using $\hpseq{p}$ is just the SVM formulation and is thus convex. Different from previous algorithms, in Alg.~\ref{alg3}, we adjust the SVM classification rate to $\eta$, which is easier to implement. Assuming we find a $C_1$ that satisfies the $\eta$-constraint using $P1$, then due to the convexity of SVM, it can be shown that the optimizing objective of P1 will be the same in both cases (exhaustive enumeration and our proposed regularization adjustment), albeit the solution $\hat{X}_p^+$ might differ (there could be multiple solutions).

\subsection{Nonlinear Extensions}
In problem $P1$, we assume a linear decision boundary generating SVMP descriptors. However, looking back at our solutions in Algorithms~\eqref{alg2} and~\eqref{alg3}, it is clear that we are dealing with standard SVM formulations to solve our relaxed objectives. In the light of this, instead of using linear hyperplanes for classification, we may use nonlinear decision boundaries by using the kernel trick to embed the data in a Hilbert space for better representation. Assuming $\dataset=\pdataset\cup\ndataset$, by the Representer theorem~\cite{smola1998learning}, it is well-known that for a kernel $K:\dataset\times \dataset\rightarrow \reals{}_+$, the decision function $f$ for the SVM problem P1 will be of the form:
\begin{equation}
f(.) = \sum_{\feat\in\pseq{}\cup\ndataset}\alpha_{\feat} K(., \feat),
\label{eq:ksvm}
\end{equation}
where $\alpha_{\feat}$ are the parameters of the non-linear decision boundaries. However, from an implementation perspective, such a direct kernelization may be problematic, as we will need to store the training set to construct the kernel. We avoid this issue by restricting our formulation to use only homogeneous kernels~\cite{vedaldi2012efficient}, as such kernels have explicit linear feature map embeddings on which a linear SVM can be trained directly. This leads to exactly the same formulations as in~\eqref{eq:mil}, except that now our features $\feat$ are obtained via a homogeneous kernel map. In the sequel, we call such a descriptor a~\emph{nonlinear SVM pooling} (NSVMP) descriptor. %The linear and non-linear SVMP descriptors are fused using a multiple-kernel learning setup for subsequent action classification.
% Thus, instead of the linear SVMP descriptors, we could use the vector of $\alpha_{\feat}$ to describe the actions in a sequence. We call such a descriptor, a~\emph{non-linear SVM pooling} (NSVMP) descriptor. 
\subsection{Temporally-Ordered Extensions}
\label{osvmp}
In the formulations we proposed above, there are no explicit constraints to enforce the temporal order of features in the SVMP descriptor. This is because, in the above formulations, we assume the features themselves capture the temporal order already. For example, the temporal stream in a two-stream model is already trained on a densely-sampled stack of consecutive optical flow frames. However, motivated by several recent works~\cite{bilen2016dynamic,grp,fernando2015modeling,dynamic_flow},
we extend our Equation~\eqref{eq:mil} by including ordering constraints as:
\begin{equation}
w^T\pfeat{i}{j} + \delta \leq w^T\pfeat{i}{k}, \quad \forall j<k; \pfeat{i}{j}, \pfeat{i}{k} \in \hpseq{i}
\end{equation} 
where we reuse the notation defined above and define $\delta>0$ as a margin enforcing the order. In the sequel, we use this temporally-ordered variant of SVMP for our video representation. Note that with the ordering constraints enforced, it is difficult to use the enumerative or alternating schemes for finding the SVMP descriptors, instead we use Alg.~\ref{alg3} by replacing the SVM solver by a custom solver~\cite{manopt}. 

\begin{figure*}[ht]
\centering
\includegraphics[width=0.75\linewidth,trim={0cm 3.5cm 0cm 2cm}, clip]{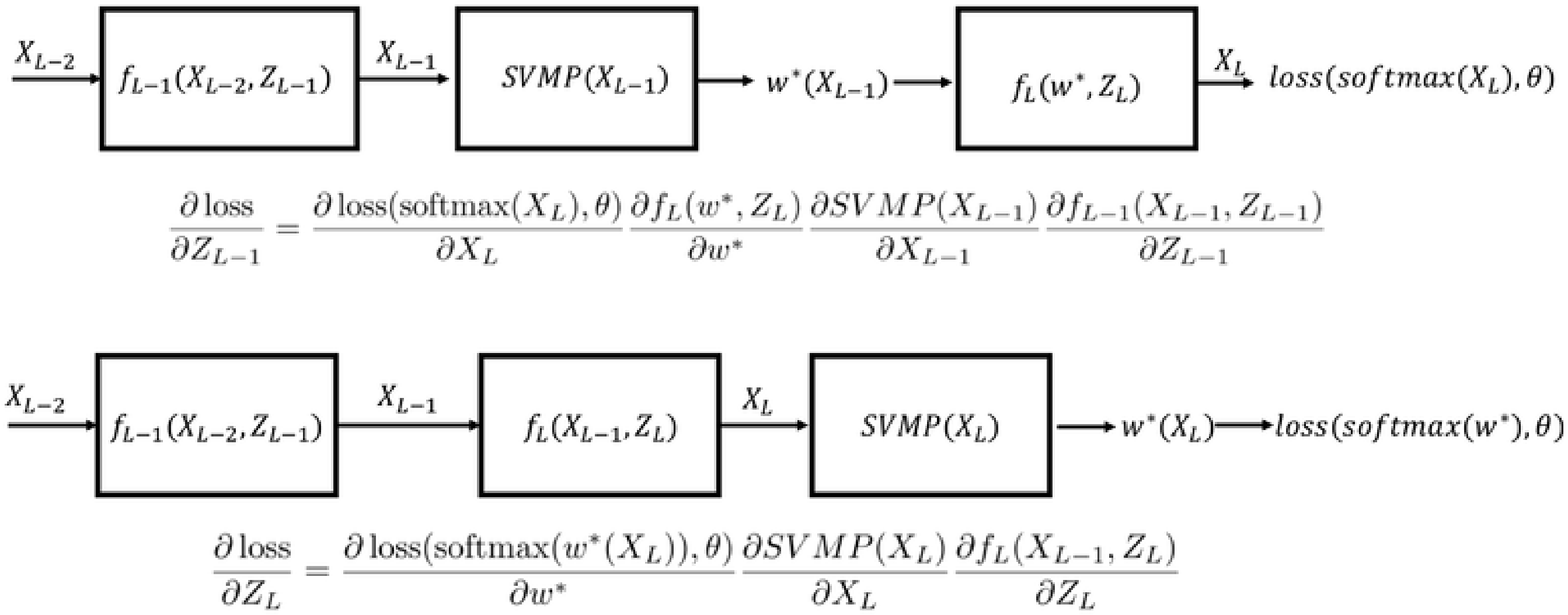}
\centering
\caption{Two possible ways to insert SVM pooling layer within a standard CNN architecture. In the first option (top), we insert the SVMP layer between fully connected layers, while in the latter we include it before the final classifier layer. The choice of $L-1$ layer for the former is arbitrary. We also show the corresponding partial gradients with respect to weights of the layer penultimate to the SVM pooling layer. Except for the gradients $\frac{\partial SVMP(X)}{\partial X}$, other gradients are the standard ones. Here $Z_{\ell}$ represents the weights of the $\ell$-th layer of the network.} 
\label{fig:dpl}
\end{figure*}
\section{End-to-End CNN Learning}
\label{sec:e2e}
In this section, we address the problem of training a CNN end-to-end with SVM pooling as an intermediate layer -- the main challenge is to derive the gradients of SVMP for efficient backpropagation. This challenge is amplified by the fact that we use the parameters of the decision hyperplane to generate our pooling descriptor, this hyperplane is obtained via a non-differentiable argmin function (refer to~\eqref{eq:mil}). However, fortunately, there is well-developed theory addressing such cases using the implicit function theorem~\cite{dontchev2009implicit}, and several recent works towards this end in the CNN setting~\cite{gould2016differentiating}. We follow these approaches and derive the gradients of SVMP below.

% In this section, we derive the necessary formulations for achieving our SVMP scheme in an end-to-end fashion within a CNN. For the sake of efficient backpropagation, we assume to know the value of $C$ in~\eqref{eq:ratio-constraint} that will satisfy the $\eta$ constraint. Ideally, the SVMP layer could be inserted between any layers in the standard CNNs. Although the formulation of gradients for backpropagation is similar, the complexity of computing the gradients varies. In this section, we will give two examples about inserting our discriminative pooling layer. Next, the gradient formulations will be derived for the specific task of discriminative pooling.

\subsection{Discriminative Pooling Layer}
In Figure~\ref{fig:dpl}, we describe two ways to insert the discriminative pooling layer into the CNN pipeline, namely (i) inserting SVMP at some intermediate layer and (ii) inserting SVMP at the end of the network just before the final classifier layer. While the latter pools smaller dimensional features, computing the gradients will be faster (as will be clear shortly). However, the last layer might only have discriminative action features alone, and might miss other spatio-temporal features that could be useful for discriminative pooling. This is inline with our observations in our experiments in Section~\ref{sec:exp} that suggest that applying discriminative pooling after pool5 or fc6 layers is significantly more useful than at the end of the fc8 layer. This choice of inserting the pooling layer between some intermediate layers of the CNN leads to the first choice. Figure~\ref{fig:dpl} also provides the gradients that need to be computed for back-propagation in either case. The only new component of this gradient is that for the argmin problem of pooling, which we derive below.

\subsection{Gradients Derivations for SVMP}
% As mentioned above, the only non-trivial component in our new SVM pooling layer is the computation of the gradient of the following function with respect to the input. Following the notation in Figure~\ref{fig:dpl}, this means computing the gradient wrt $x$ of:
% In this section, we address the problem of training a CNN end-to-end with SVM pooling as an intermediate layer -- the main challenge is to derive the gradients of SVMP for efficient backpropagation. 
Assume a CNN $f$ taking a sequence $S$ as input. Let $f_{L}$ denote the $L$-th CNN layer and let $X_{L}$ denote the feature maps generated by this layer for all frames in $S$. We assume these features go into an SVMP pooling layer and produces as output a descriptor $w$ (using a precomputed set of negative feature maps), which is then passed to subsequent CNN layers for action classification. Mathematically, let 
$g(z) = \argmin_{w} \svmp(X_{L-1})$ define the SVM pooling layer, which we re-define using hinge-loss in the objective $f(z,w)$ as: 
\begin{equation}
\svmp(X_{L-1})=\frac{1}{2}\enorm{w}^2+ \frac{\lambda}{2}\sum_{z\in X_{L-1}}\!\!\max\left(0,\theta(z;\eta)w^Tz-1\right)^2.\nonumber
\end{equation}

% This simplifies our objective to:
% \begin{align}
% &\min_Z\quad \loss(Z, g(w)) \text{ subject to }\nonumber\\
% &g(w) = \argmin_{w} f(x,w) := \nonumber\\
% &\half\enorm{w}^2 + \lambda\sum_{j} \max(0,\ \theta(x_j;\eta)\left(w_i^Tx_j\right) - 1)^2.
% \end{align}

As is by now clear, with regard to a CNN learning setup, we are dealing with a bilevel optimization problem here -- that is, optimizing for the CNN parameters via stochastic gradient descent in the outer optimization, which requires the gradient of an argmin inner optimization with respect to its optimum, i.e., we need to compute the gradient of $g(z)$ with respect to the data $z$. By applying Lemma 3.3 of~\cite{gould2016differentiating}, this gradient of the argmin at an optimum SVMP solution $w^*$ can be shown to be the following:
\begin{equation}
\nabla_{z} g(z)|_{w=w^*} = -\nabla_{ww} \svmp(X_{L-1})^{\!\!-1} \nabla_{zw} \svmp(X_{L-1}),\nonumber
\end{equation}
where the first term captures the inverse of the Hessian evaluated at $w^*$ and the second term is the second-order derivative wrt $z$ and $w$. Substituting for the components, we have the gradient at $w=w^*$ as:
{\small 
\begin{align}
%\nabla_z g(w^*) = &
-\!\!\left(\!\eye{}\!\!+\!\!\lambda\hspace*{-0.5cm}\sum_{\forall j:\theta_jw^Tz_j >1}\hspace*{-0.5cm} (\theta_jz_j)(\theta_jz_j)^T\right)^{\!\!\!\!-1}\!\!\!\left[\lambda\hspace*{-0.3cm}\sum_{\forall j: \theta_jw^Tz_j >1} \hspace*{-0.6cm}\text{D }(\theta_j^2w^Tz_j\!-\!\theta_j)\!+\!\theta_j^2wz_j^T\!\!\right]
\label{eq:bilevel}
\end{align}
}
where for brevity, we use $\theta_j = \theta(z_j; \eta)$, and $\text{D}$ is a diagonal matrix, whose $i$-th entry as $D_{ii}=\theta_i^2w^Tz_i-\theta_i$. %\footnote{As the gradient usually requires computing the inverse of a large Hessian, which may be expensive (if the DP layer uses intermediate CNN features). Thus, we use a diagonal approximation to the Hessian ($\nabla_{ww} SVMP(Y_L)$) to be used for the inverse.} See Supplementary Material for details.

\section{Experiments}
\label{sec:exp}
In this section, we explore the utility of discriminative pooling on several vision tasks, namely (i) action recognition using video and skeletal features, (ii) localizing actions in videos, (iii) image set verification, and (iv) recognizing dynamic texture videos. We introduce the respective datasets and experimental protocols in the next.
%chmark datasets used in several cutting-edge vision tasks, namely video action re We first introduce, followed by an exposition to the implementation details of our framework, analysis of the performance of each module, and extensive comparisons to previous works.
\subsection{Datasets}
\noindent\textbf{HMDB-51~\cite{kuehne2011hmdb} and UCF-101~\cite{soomro2012ucf101}:} are two popular benchmarks for video action recognition. Both datasets consist of trimmed videos downloaded from the Internet. HMDB-51 has 51 action classes and 6766 videos, while UCF-101 has 101 classes and 13320 videos. Both datasets are evaluated using 3-fold cross-validation and mean classification accuracy is reported. For these datasets, we analyze different combinations of features on multiple CNN frameworks.
% Further, this dataset also provides 27,847 textual descriptions for the videos, 66,500  temporally  localized  intervals, and 41,104 labels for 46 objects.

\noindent\textbf{Charades~\cite{sigurdsson2016hollywood}:} is an untrimmed and multi-action dataset, containing 11,848 videos split into 7985 for training, 1863 for validation, and 2,000 for testing. It has 157 action categories, with several fine-grained categories. In the classification task, we follow the evaluation protocol of ~\cite{sigurdsson2016hollywood}, using the output probability of the classifier to be the score of the sequence. In the detection task, we follow the `post-processing' protocol described in~\cite{Sigurdsson_2017_CVPR}, which uses the averaged prediction score of a small temporal window around each temporal pivot. Using the provided two-stream fc7 feature\footnote{http://vuchallenge.org/charades.html}, we evaluate the performance on both tasks using mean average precision (mAP) on the validation set. 

\noindent\textbf{Kinetics-600~\cite{kay2017kinetics}:} is one of the largest dataset for action recognition. It consists of 500K trimmed video clips over 600 action classes with at least 600 video clips in each class. Each video clip is at least 10 seconds long with a single action class label. We apply our SVMP scheme on the CNN features (2048-D) extracted from the I3D network~\cite{carreira2017quo}. 

% \noindent\textbf{MPII Cooking Activities Dataset~\cite{rohrbach2012database}:} consists of high-resolution videos captured by a static camera, showing 14 different people cooking various dishes with 64 distinct activities. There are 3,748 video clips and 1,861 clips for the background. To analyze how SVMP works with hand-crafted features, we show experiments using the publicly available 4000-D bag-of-words features computed over HOG and trajectory features from this dataset and report mean average precision (mAP) after 7-fold cross-validation. This dataset has several actions that are subtle, such as 'peeling', 'cutting in', 'cutting apart', etc., and are highly imbalanced with regard to the number of videos per action. 
% %is challenging because each activity is  to be detected and some activities are highly unbalanced compared with others. 

\noindent\textbf{MSR Action3D~\cite{li2010action} and NTU-RGBD~\cite{shahroudy2016ntu}:} are two popular action datasets providing 3D skeleton data. Specifically, MSR Action3D has 567 short sequences with 10 subjects and 20 actions, while NTU-RGBD has 56,000 videos and 60 actions performed by 40 people from 80 different view points. NTU-RGBD is by far the largest public dataset for depth-based action recognition. To analyze the performance of SVMP on non-linear features, we use a lie-algebra encoding of the skeletal data as proposed in~\cite{vemulapalli2014human} for the MSR dataset. As for NTU-RGBD, we use a temporal CNN as in~\cite{kim2017interpretable}, but uses SVMP instead of their global average pooling.
%Each subject performing each action 2-3 times and each frame in the sequence has 20 skeletal 3D joints. , containing more than 56,000 videos with 60 actions performed by 40 people from 80 different viewpoints. Each skeleton consists of the 3D coordinates of 25 body joints in each frame. %for any skeleton-based analysis.

\noindent\textbf{Public Figures Face Database (PubFig)~\cite{kumar2009attribute}:}  contains 60,000 real-life images of 200 people. All the images are collected directly from the Internet without any post-processing, which make the images in each fold have large variations in lighting, backgrounds, and camera views. Unlike video-based datasets, PubFig images are non-sequential. To generate features, we fine-tune a ResFace-101 network~\cite{masi16dowe} on this dataset and follow the evaluation protocol of~\cite{hayat2015deep}. %However, the SVMP scheme can also be used to create the descriptor in each class, which summarize multiple frame level features into one representation. 

\noindent\textbf{YUP++ dataset~\cite{feichtenhofer2017temporal}:} is recent dataset for dynamic scene understanding. It has 20 scene classes, such as Beach, Fireworks, Waterfall, Railway, etc. There are 60 videos in each class. Half of the videos are recorded by a static camera and the other half by a moving camera. Accordingly, it is divided into two sub-datasets, \textit{YUP++ moving camera} and \textit{YUP++ static camera}. We use the latest Inception-ResNet-v2 model~\cite{szegedy2017inception} to generate features (from last dense layer) from RGB frames and evaluate the performance according to the setting in~\cite{feichtenhofer2017temporal}, which use a 10/90 train-test ratio.

%, and is an extension of the previous YUPenn dataset~\cite{derpanis2012dynamic}.
\begin{figure}[htbp]
	\begin{center}
        \subfigure[]{\label{subfig:1}\includegraphics[width=0.49\linewidth,clip]{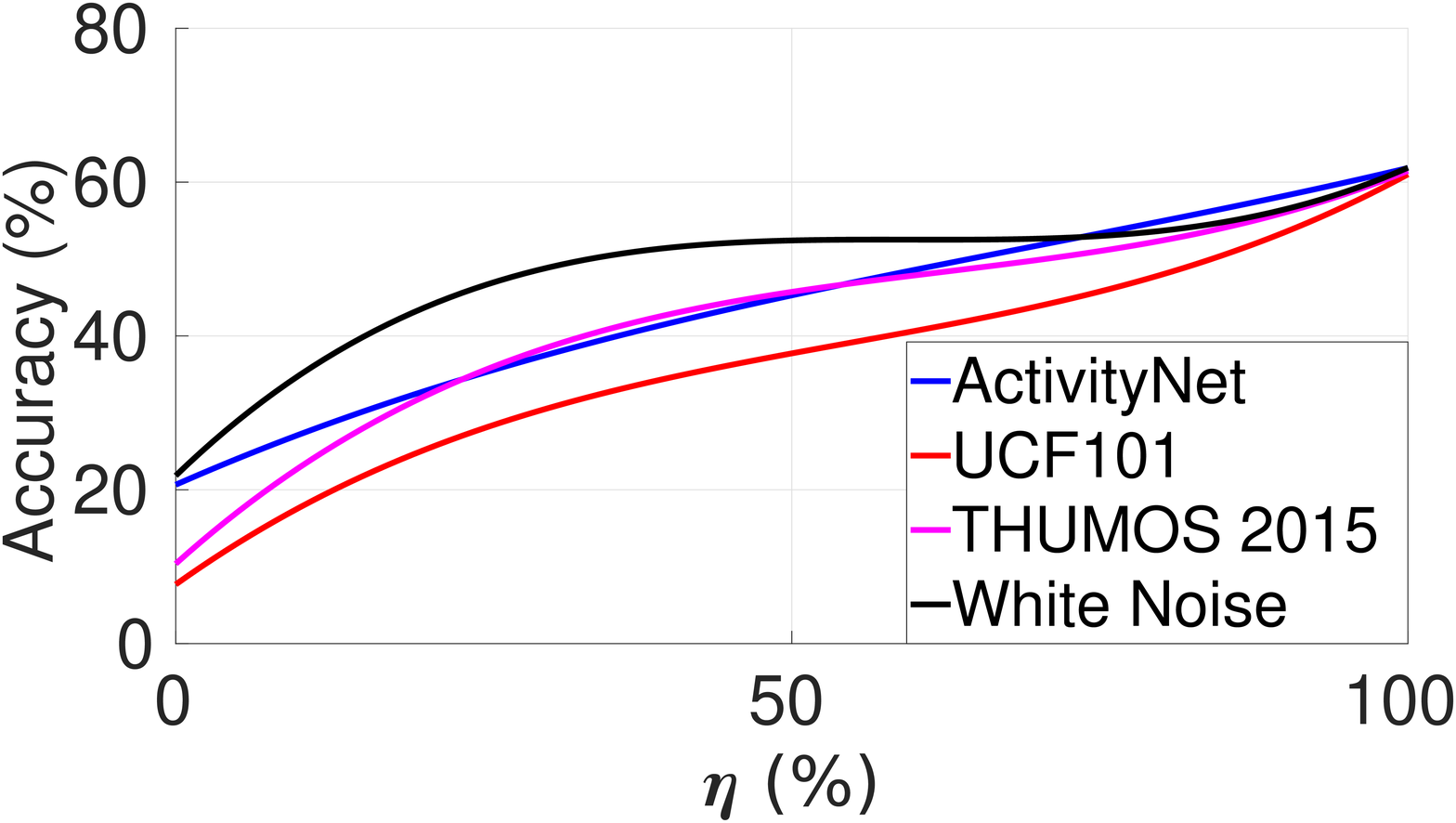}}
        \subfigure[]{\label{subfig:2}\includegraphics[width=0.49\linewidth,clip]{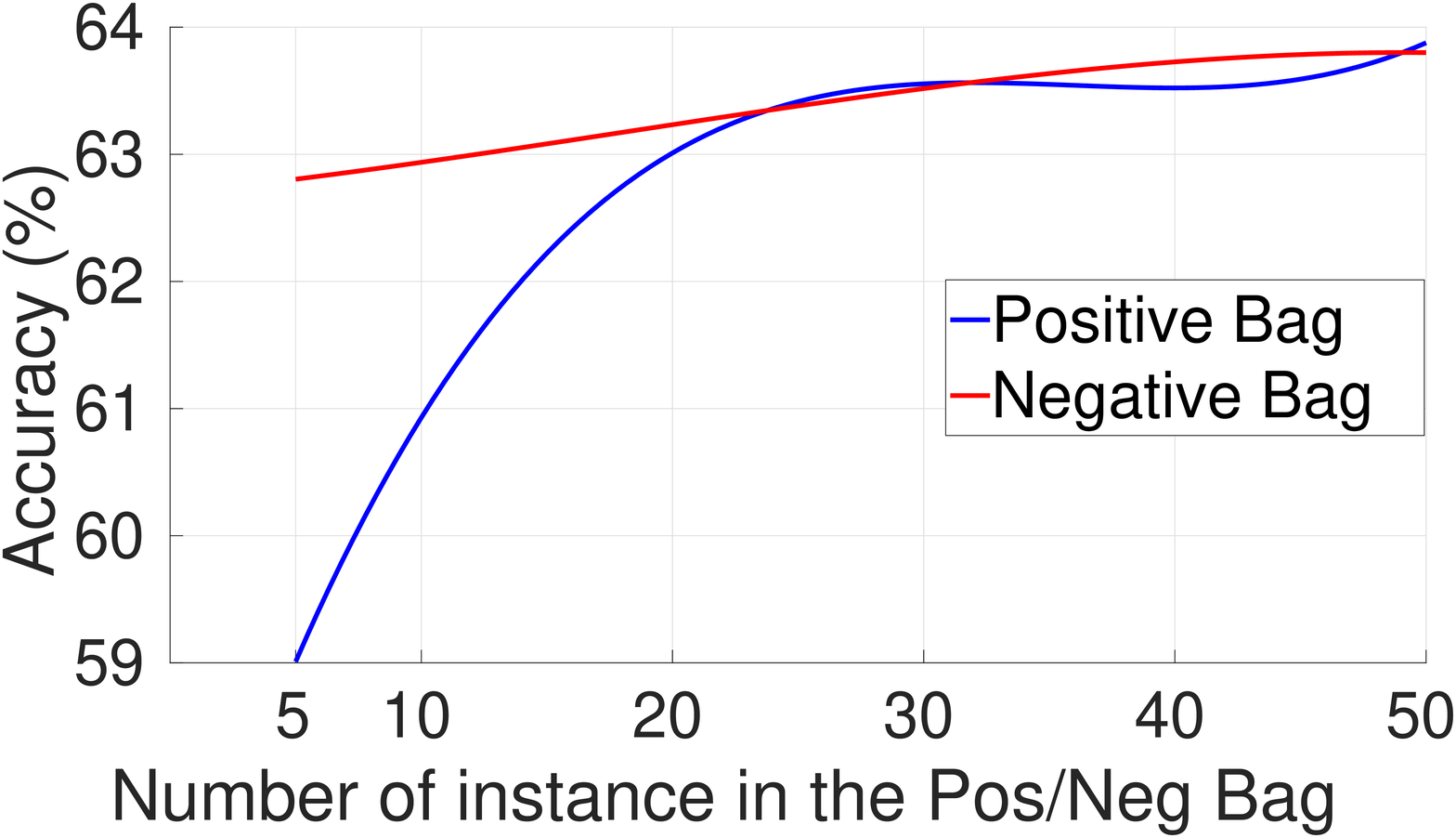}}
     \subfigure[]{\label{subfig:3}\includegraphics[width=0.49\linewidth,clip]{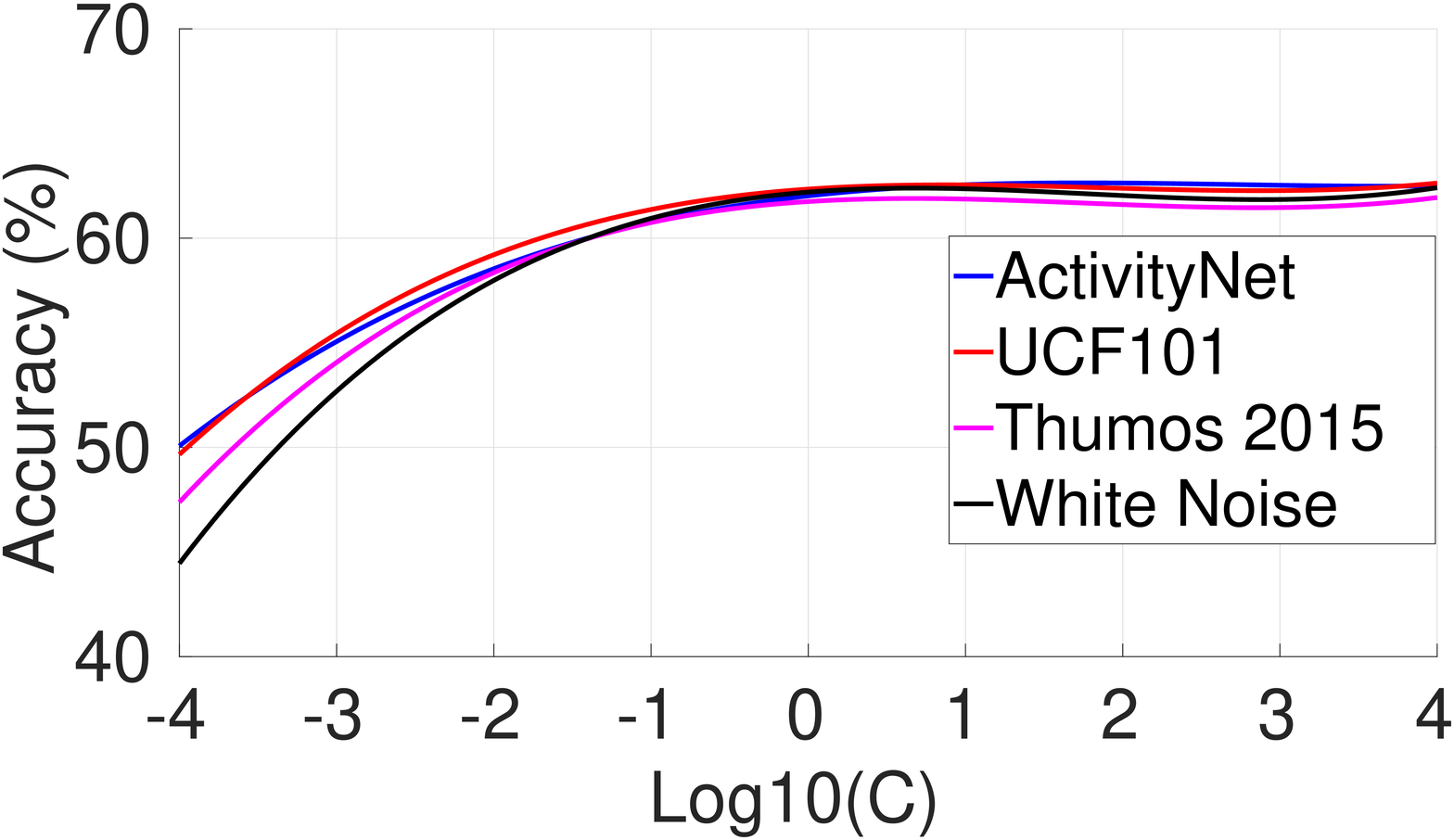}}
    \subfigure[]{\label{subfig:4}\includegraphics[width=0.49\linewidth,clip]{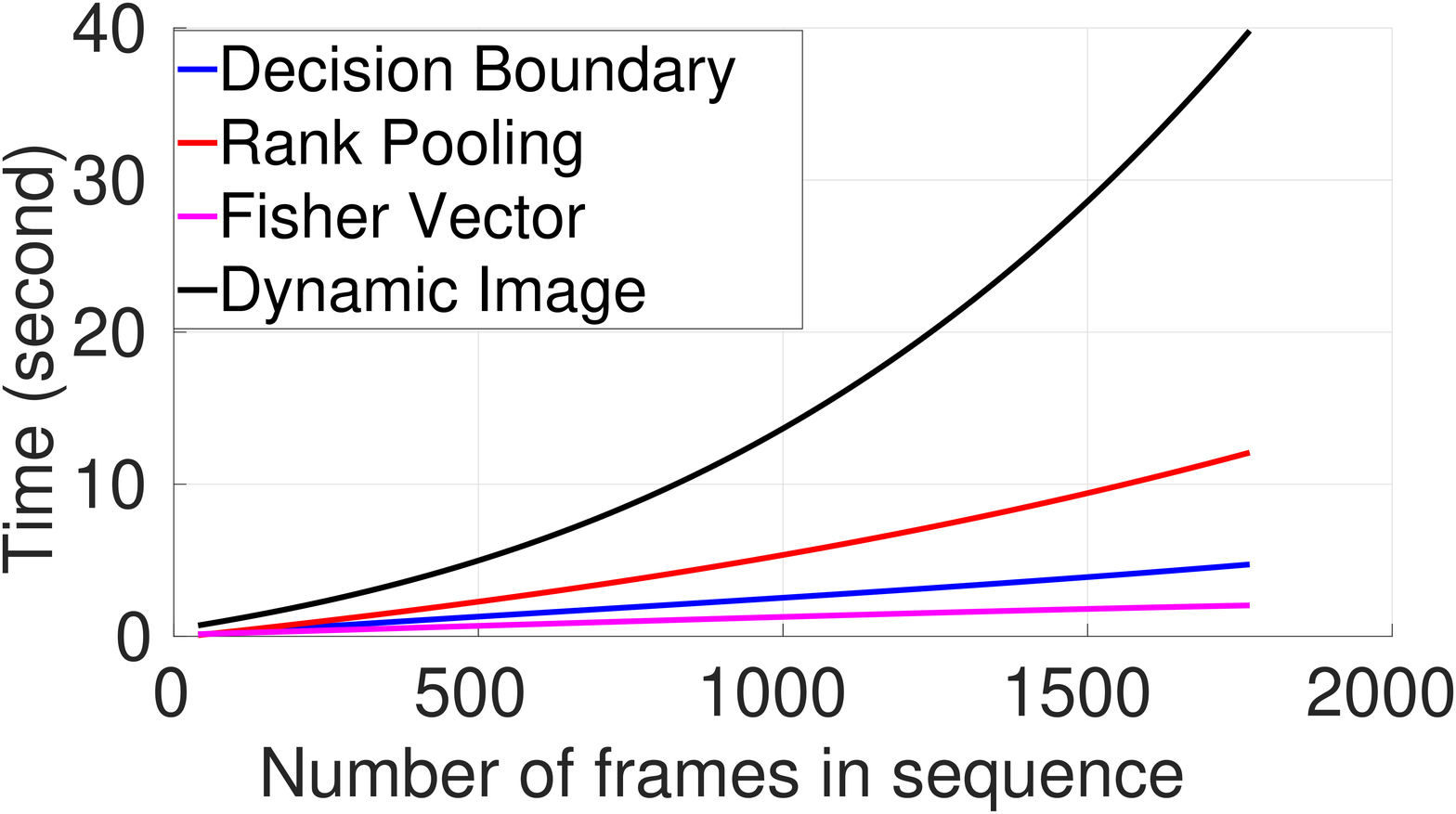}}        
	\end{center}
	\caption{Analysis of the parameters used in our scheme. All experiments use VGG features from fc6 dense layer. See text for details.}
%     See text for detailsFig.~\ref{subfig:1} shows an analysis of the influence of per-sequence accuracy threshold $\eta$ on the overall action recognition accuracy on HMDB51 split 1. Fig.\ref{subfig:2} shows the accuracy of SVMP with different number of instance in Positive/Negative bag on HMDB51 split 1. Fig.~\ref{subfig:3} shows the accuracy of SVMP descriptor against increasing slack regularization parameter $C_1$ used in our formulation~\eqref{eq:mil} on HMDB51 split 1. Fig.~\ref{subfig:4} shows the running time of each popular algorithm.
	\label{fig:all_plots}
\end{figure}

\subsection{Parameter Analysis}
In this section, we analyze the influence of each of the parameters in our scheme.
%  and apply them on the HMDB-51 dataset (viz. on UCF-101)

\noindent\textbf{Selecting Negative Bags:} An important step in our algorithm is the selection of the positive and negative bags in the MIL problem. We randomly sample the required number of frames (say, 50) from each sequence/fold in the training/testing set to define the positive bags. In terms of the negative bags, we need to select samples that are unrelated to the ones in the positive bags. We explored four different negatives in this regard to understand the impact of this selection. We compare our experiments on the HMDB-51 (and UCF101) datasets.  Our considered the following choices for the negative bgs: clips from  (ithe  ActivityNet dataset~\cite{caba2015activitynet} unrelated to HMDB-51, (ii) the UCF-101 dataset unrelated to HMDB-51, (iii) the Thumos Challenge background sequences\footnote{http://www.thumos.info/home.html}, and (iv) synthesized random white noise image sequences. For (i) and (ii), we use 50 frames each from randomly selected videos, one from every unrelated class, and for (iv) we used 50 synthesized white noise images, and randomly generated stack of optical flow images. 
%Note that for deep features in the positive bag, we pass randomly constructed images/flow images to the same CNN models and extract the feature from the same layer to make up the negative bag and for hand-crafted or geometry features, we directly use the white noise to construct negative bag, which match the dimension of features in the positive bag. 
Specifically, for the latter, we pass white noise RGB images to the same CNN models and extract the feature from the last fully-connected layer. As for hand-crafted or geometry features used in our other experiments (such as action recognition on human pose sequences), we directly use the white noise as the negative bag. As shown in Figure ~\ref{subfig:1}, the white noise negative is seen to showcase better performance for both lower and higher value of $\eta$ parameter. 

\begin{figure}[htbp]
\centering
\subfigure[Thumos]{\label{fig:thumos2}\includegraphics[width=0.3\linewidth,clip]{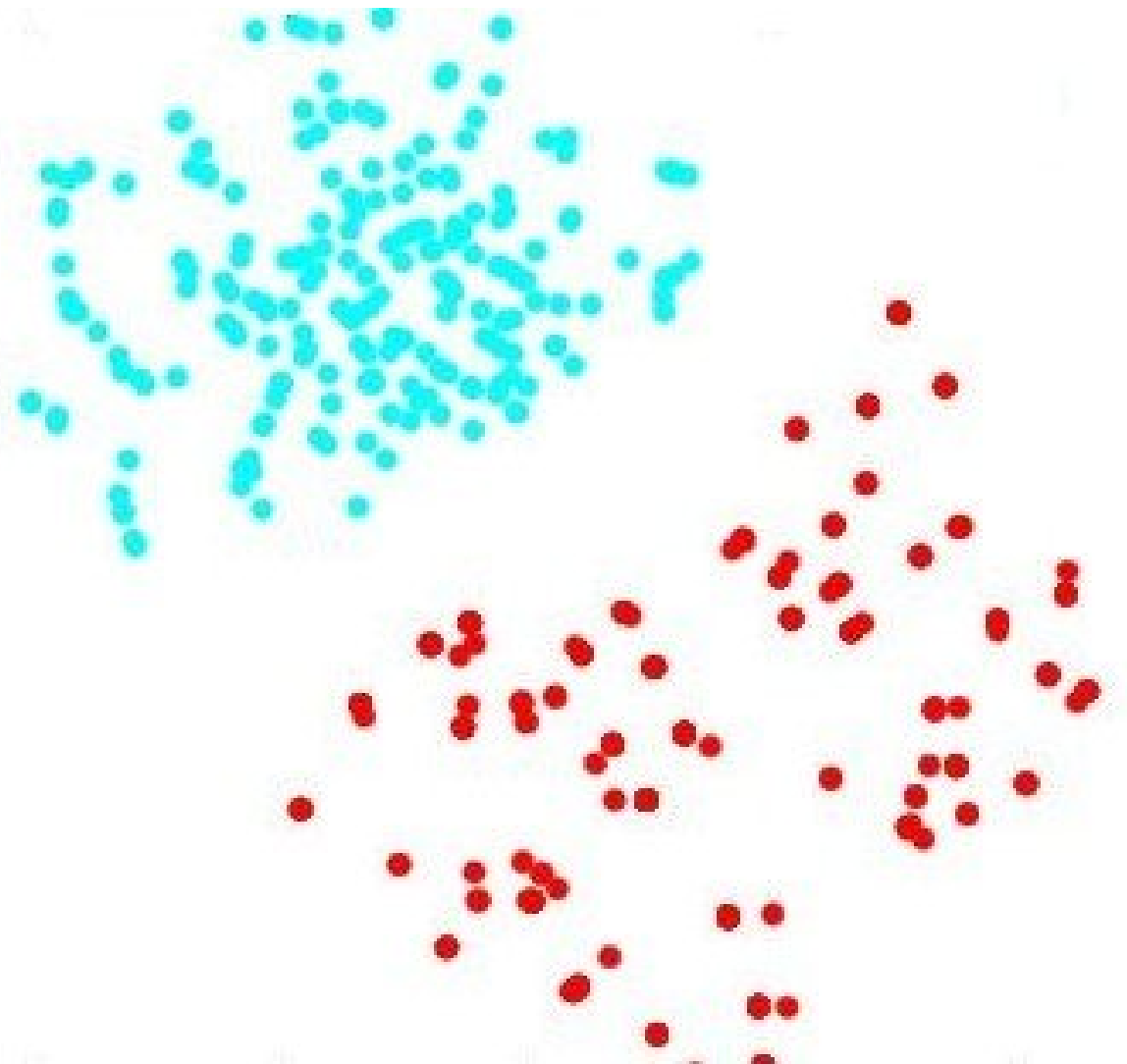}}
\subfigure[UCF101]{\label{fig:ucf2}\includegraphics[width=0.3\linewidth,clip]{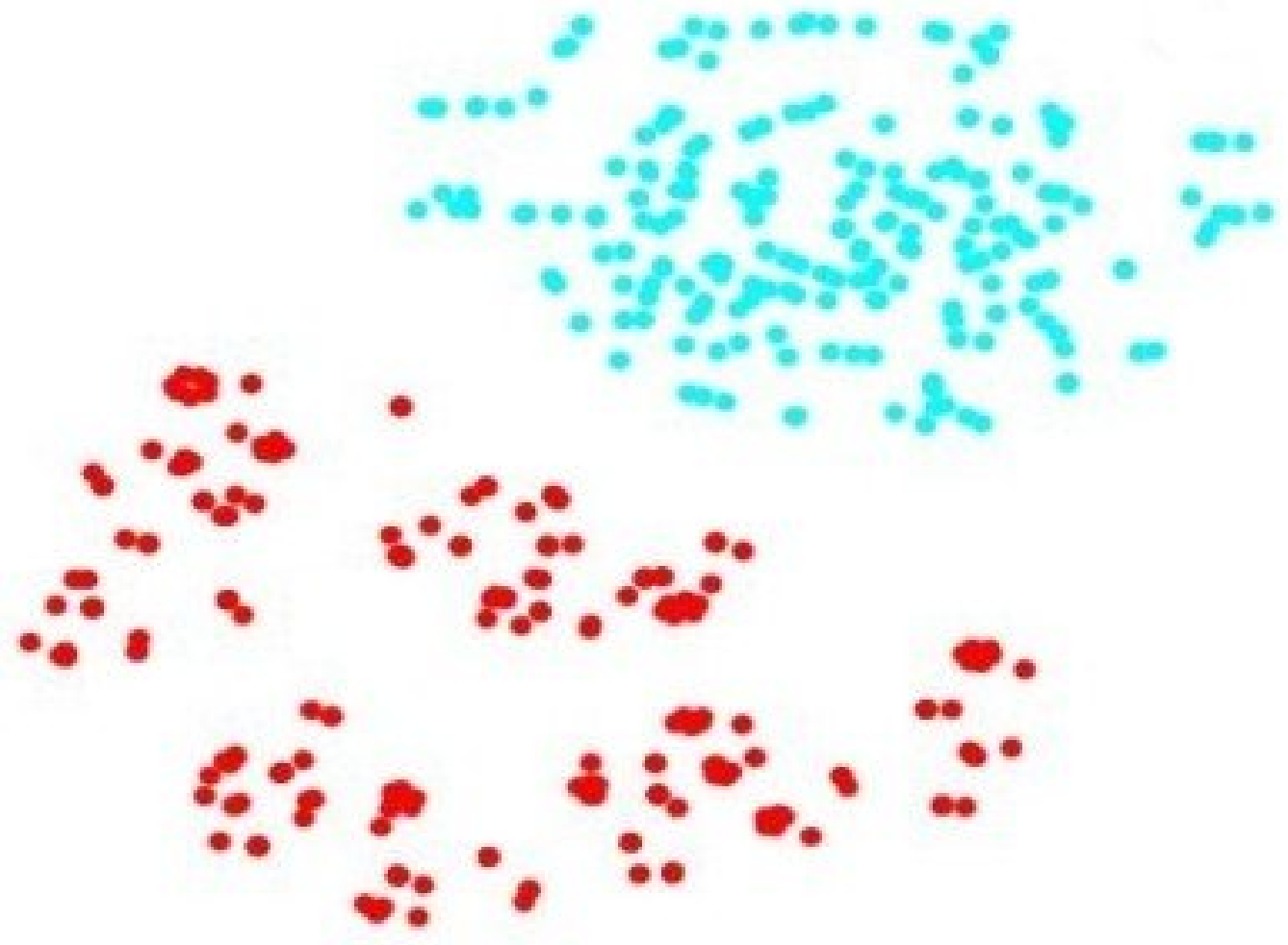}}
\subfigure[White Noise]{\label{fig:whitenoise}\includegraphics[width=0.3\linewidth,clip]{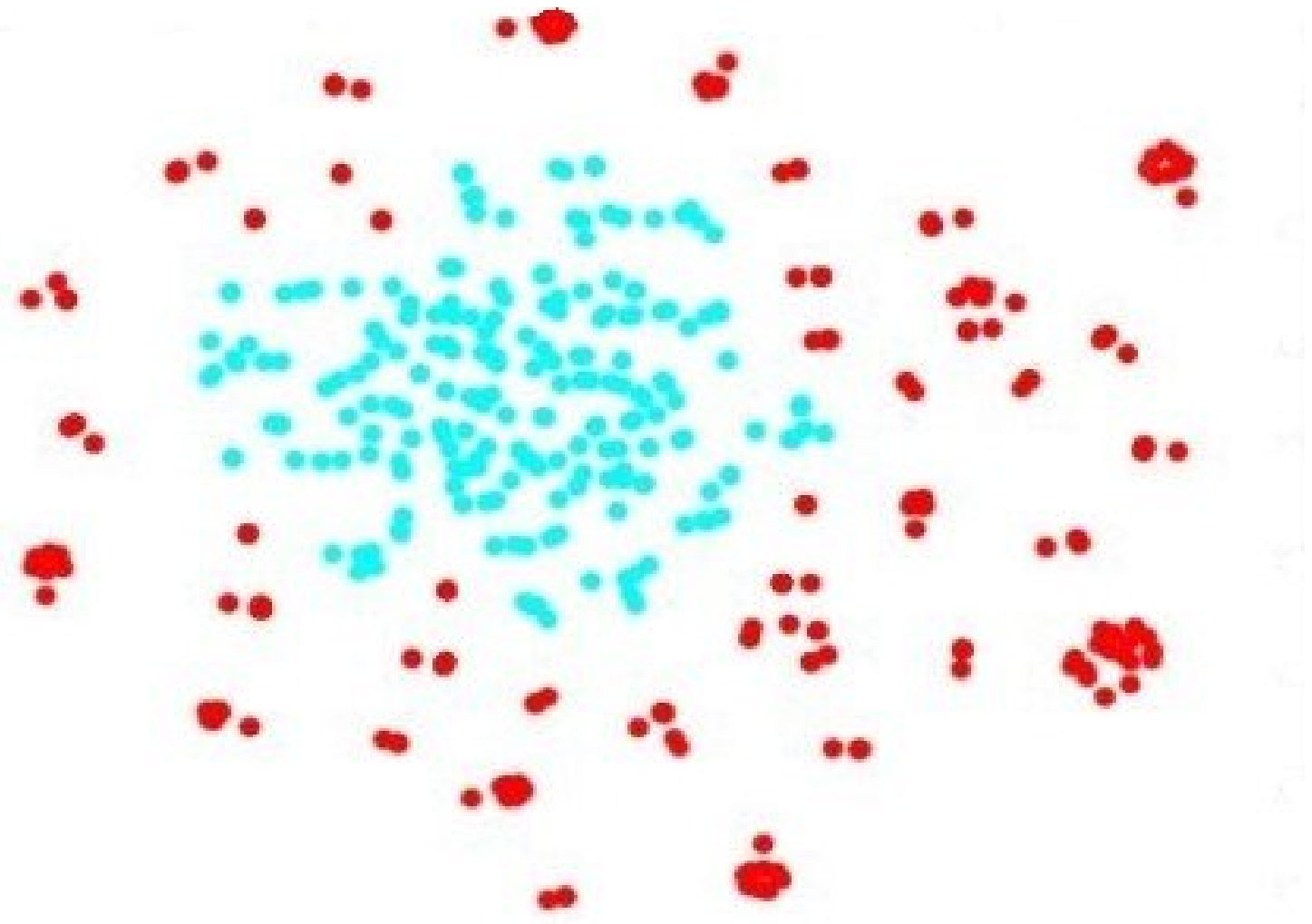}}
\caption{T-SNE plots of positive (blue) and negative bags (red) when using negatives from: (a) Thumos, (b) UCF101, and (c) white noise.}
\label{fig:tsne}
\end{figure}

To understand this trend, in Figure~\ref{fig:tsne}, we show TSNE plots visualizing the deep CNN features for the negative bag variants.  Given that the CNNs are trained on real-world image data and we extract features from the layer before the last linear layer, it is expected that these features be linearly separable (as seen in Figure~\ref{fig:thumos2} and~\ref{fig:ucf2}). However, we believe using random noise inputs may be activating combinations of filters in the CNN that are never co-activated during training, resulting in features that are  highly non-linear (as Figure~\ref{fig:whitenoise} shows). Thus, when requiring SVMP to learn linear/non-linear decision boundaries to classify video features against these ``noise'' features perhaps forces the optimizer to select those dimensions in the inputs (positive bag) that are more correlated with actions in the videos, thereby empowering the descriptor to be more useful for classification. 

% would build a difficult optimization problem to solve in finding a separating hyperplane, leading to overfitting the decision boundary to the bags, thereby creating a significantly better summarized representation.

\begin{figure}[]
	\begin{center}
        \includegraphics[width=0.9\linewidth,trim={0cm 0cm 0cm 0cm},clip]{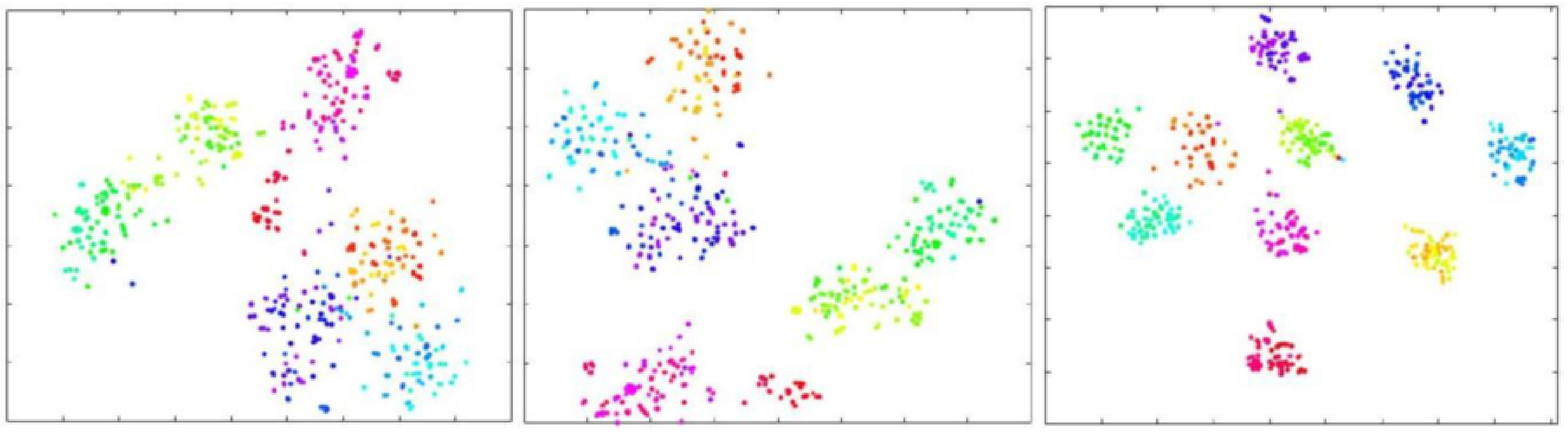}
	\end{center}
	\caption{T-SNE visualizations of SVMP and other pooling methods on sequences from the HMDB51 dataset (10 classes used). From left to right, Average Pooling, Max Pooling, and SVMP.}
   	\label{fig:4}
\end{figure}

In Figure~\ref{fig:4}, we show the TSNE visualizations of SVMP descriptors comparing to average pooling and max pooling on data from 10-classes of HDMB-51 dataset. The visualization shows that SVMP leads to better separated clusters, substantiating that SVMP is learning discriminative representations. 

\noindent\textbf{Choosing Hyperparameters:} The three important parameters in our scheme are (i) the $\eta$ deciding the quality of an SVMP descriptor, (ii) $C_1=C$ used in Algorithm~\ref{alg3} when finding SVMP per sequence, and (iii) sizes of the positive and negative bags. To study (i) and (ii), we plot in Figures~\ref{subfig:3} and~\ref{subfig:1} for HMDB-51 dataset, classification accuracy when $C$ is increased from $10^{-4}$ to $10^{4}$ in steps and when $\eta$ is increased from 0-100\% and respectively. We repeat this experiment for all the different choices of negative bags. As is clear, increasing these parameters reduces the training error, but may lead to overfitting. However, Figure~\ref{subfig:2} shows that increasing $C$ increases the accuracy of the SVMP descriptor, implying that the CNN features are already equipped with discriminative properties for action recognition. However, beyond $C=10$, a gradual decrease in performance is witnessed, suggesting overfitting to bad features in the positive bag. Thus, we use $C=10$ ( and $\eta=0.9$) in the experiments to follow. To decide the bag sizes for MIL, we plot in Figure~\ref{subfig:2}, performance against increasing size of the positive bag, while keeping the negative bag size at 50 and vice versa; i.e., for the red line in Figure \ref{subfig:2}, we fix the number of instances in the positive bag at 50; we see that the accuracy raises with the cardinality of the negative bag. A similar trend, albeit less prominent is seen when we repeat the experiment with the negative bag size, suggesting that about 30 frames per bag is sufficient to get a useful descriptor. %For our MKL setup in~\eqref{eq:mkl}, we use $\beta_1=\beta_2=1$. 

\noindent\textbf{Running Time:} In Figure~\ref{subfig:4}, we compare the time it took on average to generate SVMP descriptors for an increasing number of frames in a sequence on the UCF101 dataset. For comparison, we plot the running times for some of the recent pooling schemes such as rank pooling~\cite{bilen2016dynamic,fernando2015modeling} and the Fisher vectors~\cite{wang2013action}. The plot shows that while our scheme is slightly more expensive than standard Fisher vectors (using the VLFeat\footnote{http://www.vlfeat.org/}), it is significantly cheaper to generate SVMP descriptors than some of the recent popular pooling methods. To be comparable, we use publicly available code of SVM in SVMP as well as in rank pooling.

\subsection{Experiments on HMDB-51 and UCF-101}
Following recent trends, we use a two-stream CNN model in two popular architectures, the VGG-16 and the ResNet-152~\cite{feichtenhofer2016convolutional,simonyan2014very}. For the UCF101 dataset, we directly use publicly available models from ~\cite{feichtenhofer2016convolutional}.  For the HMDB dataset, we fine-tune a two-stream VGG/ResNet model trained for the UCF101 dataset.

\noindent\textbf{SVMP Optimization Schemes:} We proposed three different optimization strategies for solving our formulation (Section~\ref{opti_solution}). The enumerative solution is trivial and non-practical. Thus, we will only compare Algorithms~\ref{alg2} and~\ref{alg3} in terms of the performance and efficiency. In Table~\ref{algori_compari}, we show the result between the two on fc6 features from a VGG-16 model. It is clear that the alternating solution is slightly better than parameter-tuning solution; however, is also more computationally expensive. Considering the efficiency, especially for the large-scale datasets, we use parameter-tuning solution in the following experiments.

\noindent\textbf{SVMP on Different CNN Features:} We generate SVMP descriptors from different intermediate layers of the CNN models and compare their performance. Specifically, features from each layer are used as the positive bags and SVMP descriptors computed using Alg.~\ref{alg2} against the chosen set of negative bags. In Table~\ref{table:1}, we report results on split-1 of the HMDB dataset and find that the combination of fc6 and pool5 gives the best performance for the VGG-16 model, while pool5 features alone show good performance using ResNet. We thus use these feature combinations for experiments to follow. 
% we show this comparison on the split-1 of HMDB51 dataset. We evaluate the results on pool5, fc6, fc7, fc8, and the softmax output layer using the VGG-16 model and pool5 and fc1000 in the ResNet-152 model. Specifically,  The second column of Table~\ref{table:1} and \ref{table:2} shows the performance of the SVMP descriptor for each of the CNN features. As seen from the table, fc6 in VGG and pool5 in ResNet features show the best performance, better by about 6\% over the next best features. We further investigated for any complementary benefits that these features offer against other layers. In the third column in Table~\ref{table:1} and \ref{table:2}, we show the accuracy when combining features from other layers with fc6 and pool5. Interestingly, we find that fc6 combined with pool5 performs better in VGG and pool5 alone works better in ResNet. Thus, we use these combinations for the respective models in our experiments.

\begin{table}[]
\centering
\caption{Comparison between Algorithms~\ref{alg2} and~\ref{alg3} in HMDB-51 split-1.}
\label{algori_compari}
\begin{tabular}{lcc}
\hline
Method   & Accuracy   & Avg. Time (sec)/Video\\
\hline
Alternating Algorithm (Alg.~\ref{alg2}) & \textbf{69.8\%} & 2.4\\
Parameter-tuning Algorithm (Alg.~\ref{alg3})    & 69.5\%    &\textbf{0.2}                               
\end{tabular}
\end{table}

\begin{table}[]
\centering
\caption{Comparison of SVMP descriptors using various CNN Features on HMDB split-1.}
\label{table:1}
\begin{tabular}{lcc}
\hline
%Layer   & \multicolumn{2}{l}{Accuracy} \\\hline
Feature/   & Accuracy   & Accuracy when\\
model        & independently & combined with:\\
\hline
pool5 (vgg-16)   & 57.9\%         & \textbf{63.8\%} (fc6)                                  \\
fc6 (vgg-16)     & 63.3\%      & -                              \\
fc7 (vgg-16)     & 56.1\%         & 57.1\% (fc6)                                 \\
fc8 (vgg-16)     & 52.4\%         & 58.6\% (fc6)                                 \\
softmax (vgg-16) & 41.0\%         & 46.2\% (fc6)   \\
\hline
pool5 (ResNet-152)  & \textbf{69.5\%}         & -                               \\
fc1000 (ResNet-152)    & 61.1\%    & 68.8\% (pool5)
\end{tabular}
\end{table}

\noindent\textbf{Linear vs Non-Linear SVMP:} We analyze the complementary nature of SVMP and its non-linear extension NSVMP (using a homogeneous kernel) on HMDB-51 and UCF-101 split1. The results are provided in Table~\ref{table:3}, and clearly show that the combination leads to significant improvements consistently on both datasets. 

\noindent\textbf{End-to-End Learning and Ordered-SVMP:} In Table~\ref{table:4}\footnote{All experiments in Table~\ref{table:4} use the same input features.}, we compare to the end-to-end learning setting as described in Section~\ref{sec:e2e}. For end-to-end learning, we insert our discriminative pooling layer after the 'fc6' layer in VGG-16 model and the 'pool5' layer in ResNet model. We also present results when using the temporal ordering constraint (TC) into the SVMP formulation to build the ordered-SVMP. From the results, it appears that although the soft-attention scheme performs better than average pooling, it is inferior to SVMP itself; which is unsurprising given it does not use a max-margin optimization. Further, our end-to-end SVMP layer is able to achieve similar (but slightly inferior) performance to SVMP, which perhaps is due to the need to approximate the Hessian. As the table shows, we found that the temporal ranking is indeed useful for improving the performance of na\"ive SVMP. Thus, in the following experiments, we use SVMP with temporal ranking for all video-based tasks.
% As mentioned in the Equation~\ref{eq:ksvm} and ~\ref{eq:mkl}, non-linear decision boundary is a variant of the linear one, which is able to represent the non-linear feature. As is shown in the Table~\ref{table:3}, we concatenate the feature from selected layer in spatial and temporal stream, and apply both linear and non-linear SVMP. It is obvious that the non-linear SVMP descriptor is complementary to the linear one. Thus, we will use the combination, namely SVMP, for the following experiments.
\begin{table}[]
\centering
\caption{Comparison between SVMP and NSVMP on split-1.}
\label{table:3}
\begin{tabular}{l|ll|ll}
\hline
 \multicolumn{1}{l|}{} & \multicolumn{2}{c|}{HMDB-51} & \multicolumn{2}{c}{UCF-101} \\ \hline
       & VGG & ResNet &VGG & ResNet \\\hline
linear-SVMP   & 63.8\% & 69.5\% &91.6\% & 92.2\% \\
nonlinear-SVMP  & 64.4\% & 69.8\% &92.0\% & 93.1\% \\
Combination & \textbf{66.1\%}&  \textbf{71.0\%} &\textbf{92.2\%} &\textbf{94.0\%}    
\end{tabular}
\end{table}

% \noindent\textbf{Experiments on Hand-crafted Features: } As alluded to above, we use the MPII Cooking activities dataset for analyzing the influence of SVMP on hand-crafted features. We use the 4000-D bag-of-words encoded HOG and dense trajctories as our pooling features. In Table~\ref{table:6}, we report results from this experiment. As is clear, our scheme shows superior performance against the best prior methods (such as rank pooling~\cite{fernando2015modeling}) on this dataset, showing that our scheme is agnostic to feature type. 
% For this experiment, we use the publicly available hand-crafted features (HOG and trajectories) along with the dataset. These features are encoded in a bag-of-words setup using 4000 code words, i.e., each video frame is encoded by a 4000-D feature vectors, each for HOG and trajectories separately\footnote{There are other hand-crafted features as well, but we thought to report our analysis to HOG and trajectories due to space constraints.}. These features are then encoded using SVMP for every clip and then classified for actions.

\begin{table}[ht]
\centering
\caption{Comparison to standard pooling methods on split-1. TC is short for Temporal Constraint, E2E is short for end-to-end learning.}
\label{table:4}
\begin{tabular}{l|ll|ll}
\hline
 \multicolumn{1}{l|}{} & \multicolumn{2}{c|}{HMDB-51} & \multicolumn{2}{c}{UCF-101} \\ \hline
       & VGG & ResNet &VGG & ResNet \\\hline
Spatial Stream-AP\cite{feichtenhofer2016spatiotemporal,feichtenhofer2016convolutional}   
	   & 47.1\% & 46.7\% &82.6\% & 83.4\% \\
%Spatial Stream-MP  
%	   & 46.5\% & 45.1\% &82.2\% & 82.7\% \\
% Spatial Stream-Attention  
% 	   & 54.2\% & 51.5\% &80.3\% &81.2\% \\
Spatial Stream-SVMP   
	   & 58.3\% & 57.4\% &85.7\% &87.6\% \\
Spatial Stream-SVMP(E2E)   
	   & 56.4\% & 55.1\% &83.2\% &85.7\% \\
Spatial Stream-SVMP+TC  
	   & \textbf{59.4\%} & \textbf{57.9\%} &\textbf{86.6\%} & \textbf{88.9\%} \\
       \hline
Temporal Stream-AP \cite{feichtenhofer2016spatiotemporal,feichtenhofer2016convolutional} 
       & 55.2\% & 60.0\% &86.3\% & 87.2\% \\
%Temporal Stream-MP
%       & 54.8\% & 58.5\% &86.5\% & 86.1\% \\
% Temporal Stream-Attention
%        & 56.7\% & 62.6\% &86.9\% & 87.6\% \\
Temporal Stream-SVMP
       & 61.8\% &65.7\% &88.2\% & 89.8\% \\
Temporal Stream-SVMP(E2E)
       & 58.3\% &63.2\% &87.1\% & 87.8\% \\
Temporal Stream-SVMP+TC
       & \textbf{62.6\%} &\textbf{67.1\%} &\textbf{88.8\%} & \textbf{90.9\%} \\
       \hline
Two-Stream-AP \cite{feichtenhofer2016spatiotemporal,feichtenhofer2016convolutional} 
       & 58.2\%& 63.8\% &90.6\% &91.8\% \\
%Two-Stream-MP
%       & 56.7\% & 60.6\% &90.1\% &87.4\% \\
% Two-Stream-Attention
%        & 62.2\% & 66.5\% &90.8\% &92.5\% \\
Two-Stream-SVMP
       &66.1\% &71.0\%  &92.2\% &94.2\% \\
Two-Stream-SVMP(E2E)
       &63.5\% &68.4\%  &90.6\% &92.3\% \\
Two-Stream-SVMP+TC
       &\textbf{67.2\%}&\textbf{71.3\%}  &\textbf{92.5\%} &\textbf{94.8\%}
\end{tabular}
\end{table}

\begin{figure}[]
	\begin{center}
        \includegraphics[width=0.75\linewidth,trim={0cm 0cm 0cm 0cm},clip]{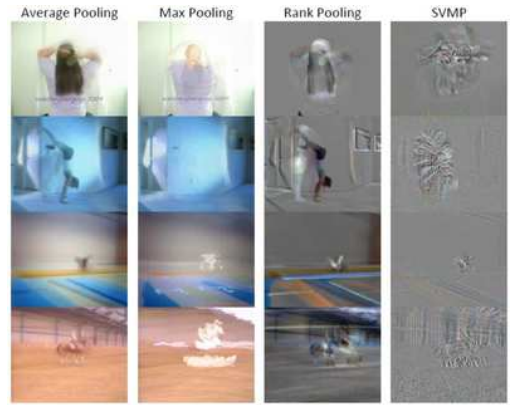}
	\end{center}
	\caption{Visualizations of various pooled descriptors.}
   	\label{fig:5}
\end{figure}

\noindent\textbf{SVMP Image:} In Figure~\ref{fig:5}, we visualize SVMP descriptor when applied directly on raw video frames. We compare the resulting image against those from other schemes such as the dynamic images of~\cite{bilen2016dynamic}. It is clear that SVMP captures the essence of action dynamics in more detail. To understand the action information present in these images, we trained an action classifier directly on these images, as is done on Dynamic images in~\cite{bilen2016dynamic}. We use the BVLC CaffeNet~\cite{jia2014caffe} as the CNN -- same the one used in~\cite{bilen2016dynamic}. The results are shown in the Table~\ref{svmpimage} on split-1 of JHMDB (a subset of HMDB-51, containing 21 classes) and UCF-101. As is clear, SVMP images are seen to outperform \cite{bilen2016dynamic} by a significant margin, suggesting that SVMP captures more discriminative and useful action-related features. Howeer, we note that in contrast to dynamic images, our SVMP images do not intuitively look like motion images; this is perhaps because our scheme captures different information related to the actions, and we do not use smoothing (via running average) when generating them. The use of random noise features as the negative bag may be adding additional artifacts. %We also note that one could apply the SVMP images for higher-layers of the network, as has been attempted for rank pooling~\cite{fernando2016discriminative}; however, we find that our scheme outperforms such attempts as the comparisons show in Table~\ref{table:10}.

%Note that achieving state-of-the-art performance is not the goal of this experiment; instead the goal is to quantitatively analyze if SVMP images carry more details of actions than Dynamic Images.

\begin{table}[]
\centering
\caption{Recognition rates on split-1 of JHMDB and UCF-101.}
\label{svmpimage}
\begin{tabular}{lcc}
\hline
Datasets   & JHMDB   & UCF-101\\
\hline
Mean image&31.3\% &52.6\%\\
Max image &28.6\% &48.0\%\\
Dynamic image~\cite{bilen2016dynamic}  & 35.8\% & 57.2\%\\
SVMP image    & \textbf{45.8\%}    &\textbf{65.4\%}                             
\end{tabular}
\end{table}

\subsection{Action Recognition at Large Scale}
Kinetics-600 is one the largest state-of-the-art dataset for action recognition on trimmed videos. For this experiment, we use the I3D network~\cite{carreira2017quo} (using the Inception-V3 architecture), as the baseline for feature generator. This model is pre-trained on ImageNet dataset~\cite{krizhevsky2012imagenet} and stacks 64 continuous frames as inputs. Specifically, we extract the CNN features from the second last layer (Mix5c) and apply average pooling to reshape the feature from 4 x 7 x 7 x 1024 into 1024-D vector for each 64-chunk of RGB frames. For each video clip, we use a sliding window to generate a sequence of such features with a window size of 64 and a temporal stride of 8 frames. Then, we apply our proposed SVMP to generate video descriptors for action recognition. In Table~\ref{kinectic}, we make comparisons with the baseline result on the validation set of Kinetics-600, and indicates that SVMP can bring clear improvements even on the large-scale setting. 
\begin{table}[]
\centering
\caption{Comparisons on Kinetics-600 dataset using I3D feature.}
\label{kinectic}
\begin{tabular}{lc}
\hline
Method  						& Accuracy\\\hline
AP~\cite{carreira2018short}		 		& 71.9\% \\
MP &67.8\% \\
SVMP   &\textbf{73.5\%}\\                                     
\end{tabular}
\end{table}
% the standard evaluation setting is to use the prediction score of 25 equidistant temporal points in the sequence, which is not suitable for any pooling scheme. Thus,
\subsection{Action Recognition/Detection in untrimmed videos} We ues the  Charades untrimmed dataset for this task. We use the publicly available two-stream VGG features from the fc7 layer for this dataset. We trained our models on the provided training set (7985 videos), and report results (mAP) on the provided validation set (1863 videos) for the tasks of action classification and detection. In the classification task, we concatenate the two-stream features and apply a sliding window pooling scheme to create multiple descriptors. Following the evaluation protocol in~\cite{sigurdsson2016hollywood}, we use the output probability of the classifier to be the score of the sequence. In the detection task, we consider the evaluation method with post-processing proposed in~\cite{Sigurdsson_2017_CVPR}, which uses the averaged prediction score of a temporal window around each temporal pivots. Instead of average pooling, we apply the SVMP. From Table~\ref{table:5}, it is clear that SVMP improves performance against other pooling schemes by a significant margin; the reason for this is perhaps the following. During training, we use trimmed video clips, however, when testing, we extract features from every frame/clip in the untrimmed test video. As the network has seen only action-related frames during training, features from background frames may result in arbitrary predictions; and average pooling or max pooling on those features would hurt performance. When optimizing the binary classification problem between positive and negative bags for SVMP, the decision boundary would capture the most discriminative data support, leading to better summary of the useful features and leading to improved performance.

% Unlike trimmed video sequences that are already clipped according to the ground truth label, untrimmed video usually contain more background or noise frames. When applying the SVMP under this scenario, the SVMP descriptor would capture more discriminative feature compared with other pooling schemes. Also, we notice that the size of sliding window should be larger than 25 frames, otherwise the performance will drop. This might because learning the decision boundary in our scheme require a minimum number of points to characterize the data distribution. 
\begin{table}[]
\centering
\caption{Comparisons on Charades dataset.}
\label{table:5}
\begin{tabular}{lccc}
\hline
Tasks  						& AP   & MP & SVMP\\\hline
Classification (mAP)		 		& 14.2\% & 15.3  & \textbf{26.3\%} \\
Detection (mAP)   & 10.9\% & 9.2 &\textbf{15.1\%}\\                                     
\end{tabular}
\end{table}
\subsection{SVMP Evaluation on Other Tasks}
In this section, we provide comprehensive evaluations justifying the usefulness of SVMP on non-video datasets and non-action tasks. We consider experiments on images sets recognition, skeleton-sequence based action recognition, and dynamic texture understanding.
%Till now, we have observed the promising performance in HMDB-51, UCF-101, Kinetics-600 and Charades datasets. All of them are image based video datasets for action recognition or detection, and our SVMP scheme is operated on the top of deep CNN features. 
%In this section, we evaluate SVMP for tasks not invl of SVMP by showing experiments in several different datasets. %Compared with the baseline result, we achieve better performance in these datasets, from $3\%$ to $14\%$. And the comparison with state-of-the-art algorithms will be given in the next section.

\textbf{MSR Action3D:} In this experiment, we explore the usefulness of SVMP on non-linear geometric features. Specifically,  we chose the scheme of Vemulapilli et al.~\cite{vemulapalli2014human} as the baseline that generates Lie algebra based skeleton encodings for action recognition. While they resort to a dynamic time warping kernel for the subsequent encoded skeleton pooling, we propose to use SVMP instead. We use the random noise with the dataset mean and deviation as the negative bag, which achieve better performance. 

\textbf{NTU-RGBD:} On this dataset, we apply our SVMP scheme on the skeleton-based CNN features.  Specifically, we use~\cite{kim2017interpretable} as the baseline, which applies a temporal CNN with residual connections on the vectorized 3D skeleton data. We swap the global average pooling layer in~\cite{kim2017interpretable} by SVM pooling layer. For the evaluation, we adopt the official cross-view and cross-subject protocols. What's interesting here is we try to explore whether the dimension of the feature point would affect the SVMP performance. During the SVMP, we use feature points with dimension from 150 to 4096. It seems only the number of data points would affect the performance of SVMP (from Charades dataset experiment), and it is not sensitive for the dimensionality.

\textbf{PubFig:} In this task, we evaluate the use of SVMP for image set representation. We follow the evaluation setting in~\cite{hayat2015deep} and create the descriptor for the training and testing by applying SVMP over ResFace-101~\cite{masi16dowe} features from every image in the PubFig dataset. Unlike the video-based tasks, all input features in this setting are useful and represent the same person; however their styles vary significantly, which implies the CNN features may be very different even if they are from the same person. This further demands that SVM pooling would need to find discriminative dimensions in the features that are correlated and invariant to the person identity.

%This requires the pooling scheme characterizing the feature mainly from the dimension level, that is to decide which dimension in the feature are more useful to represent the person.

\textbf{YUP++:} To investigate our SVMP scheme on deeper architectures, we use features from the latest Inception-ResNet-v2 model~\cite{szegedy2017inception}, which has achieved the state-of-the-art performance on the 2015 ILSVRC challenge. Specifically, we extract the RGB frames from videos and divide them into training and testing split according to the setting in~\cite{feichtenhofer2017temporal} (using a 10/90 train test ratio). Like the standard image-based CNNs, the clip level label is used to train the network on every frame.

\begin{table}[tbp]
\centering
\caption{Accuracy comparison on different subsets of HMDB-51(H) and UCF-101(U) split-1 using I3D+ features.}
\label{tab:11}
\begin{tabular}{l|l|l|l|l|l|l}
\hline
Min \# of frames    & 1               & 80                         & 140                    & 180             & 260             \\ \hline
\# of classes (H) & 51              & 49                         & 27                        & 21              & 12              \\ \hline
\# of classes (U)  & 101             & 101                       & 95                      & 82              & 52              \\ \hline
I3D (H)           & 79.6\%          & 81.8\%                & 84.1\%                 & 78.0\%          & 77.3\%          \\ \hline
SVMP (H)          & \textbf{80.0\%} & \textbf{82.9\%}  & \textbf{84.8\%} & \textbf{85.1\%} & \textbf{86.8\%} \\ \hline
I3D (U)            & 98.0\%          & 98.0\%                  & 98.0\%                 & 95.9\%          & 93.8\%          \\ \hline
SVMP (U)           & \textbf{98.4\%} & \textbf{98.9\%}  & \textbf{99.3\%} & \textbf{98.5\%} & \textbf{97.3\%} \\ \hline
\end{tabular}
\end{table}

\begin{table}[]
\centering
\caption{Comparison to the state of the art in each dataset, following the official evaluation protocol for each dataset.}
\label{table:10}
\scalebox{0.95}{
\begin{tabular}{@{}l|c|c@{}}\hline
\multicolumn{3}{c}{HMDB-51 \& UCF-101 (accuracy over 3 splits)}                      \\\hline
Method                       & HMDB-51        & UCF-101     \\\hline
Temporal segment networks\cite{Wang2016}    & 69.4\%         & 94.2\%      \\
AdaScan\cite{Kar_2017_CVPR}           & 54.9\%         & 89.4\%      \\
AdaScan + IDT + C3D\cite{Kar_2017_CVPR}           & 66.9\%         & 93.2\%      \\
ST ResNet\cite{feichtenhofer2016spatiotemporal}                    & 66.4\%         & 93.4\%      \\
ST ResNet + IDT\cite{feichtenhofer2016spatiotemporal}                    & 70.3\%         & 94.6\%      \\
ST Multiplier Network\cite{feichtenhofer2017spatiotemporal}        & 68.9\%         & 94.2\%      \\
ST Multiplier Network + IDT\cite{feichtenhofer2017spatiotemporal}        & 72.2\%         & 94.9\%      \\
Hierarchical rank pooling\cite{fernando2016discriminative} &65.0\% &90.7\% \\
Two-stream I3D\cite{carreira2017quo}               & 66.4\%        & 93.4\%      \\
Two-stream I3D+ (Kinetics 300k)\cite{carreira2017quo}               & 80.7\%        & 98.0\% \\\hline
Ours (SVMP)                  & 71.3\%         & 94.6\%      \\
Ours (SVMP+IDT)              & \textbf{72.6}\%         & \textbf{95.0}\%     \\
Ours (I3D+)              & \textbf{81.8}\%         & \textbf{98.5}\%      \\\hline
\multicolumn{3}{c}{Kinetics-600}                                \\\hline
Method                       & \multicolumn{2}{c}{Accuracy} \\\hline
I3D RGB\cite{carreira2018short}   & \multicolumn{2}{c}{71.3\%}   \\
Second-order Pooling~\cite{cherian2017second} &  \multicolumn{2}{c}{54.7\%} \\\hline
Ours (SVMP)                  & \multicolumn{2}{c}{\textbf{73.5\%}}   \\\hline
\multicolumn{3}{c}{Charades (mAP)}                                \\\hline
Method                       & Classification         & Detection        \\\hline
Two-stream\cite{simonyan2013deep}   & 14.3\%       & 10.9\%       \\
ActionVlad + IDT\cite{girdhar2017actionvlad}             & 21.0\%           & -    \\
%Two-stream + LSTM\cite{Sigurdsson_2017_CVPR}   & 17.8\%       & 10.4\%       \\
%Two-stream Extended\cite{Sigurdsson_2017_CVPR}   & 18.6\%       & 11.6\%       \\
Asynchronous Temporal Fields~\cite{Sigurdsson_2017_CVPR} & 22.4\%           & 12.8\%       \\\hline
Ours (SVMP)                  & 26.3\%           & 15.1\%    \\
Ours (SVMP+IDT)              & \textbf{27.4\%}  & \textbf{16.3\%} \\\hline
% \multicolumn{3}{c}{MPII Cooking Activity}                   \\\hline
% Method                       & \multicolumn{2}{c}{Accuracy} \\\hline
%                              & \multicolumn{2}{c}{}         \\
%                              & \multicolumn{2}{c}{}         \\\hline
% Ours (SVMP)                  & \multicolumn{2}{c}{\textbf{55.0\%}}         \\\hline
\multicolumn{3}{c}{MSR-Action3D}                            \\\hline
Method                       & \multicolumn{2}{c}{Accuracy} \\\hline
Lie Group\cite{vemulapalli2014human}               & \multicolumn{2}{c}{92.5\%}   \\
ST-LSTM + Trust Gate\cite{liu2017skeleton}    & \multicolumn{2}{c}{94.8\%}   \\\hline
Ours (SVMP)                  & \multicolumn{2}{c}{\textbf{95.5\%}}   \\\hline
\multicolumn{3}{c}{NTU-RGBD}                                \\\hline
Method                       & Cross-Subject  & Cross-View  \\\hline
Res-TCN\cite{kim2017interpretable}                 & 74.3\%         & 83.1\%      \\
ST-LSTM + Trust Gate\cite{liu2017skeleton}    & 69.2\%         & 77.7\%      \\\hline
Ours (SVMP)                  & \textbf{79.4\%}         & \textbf{87.6\%}      \\\hline
\multicolumn{3}{c}{PubFig}                                  \\\hline
Method                       & \multicolumn{2}{c}{Accuracy}\\\hline
Deep Reconstruction Models\cite{hayat2015deep}   & \multicolumn{2}{c}{89.9\%}   \\
ESBC\cite{hayat2017empowering}                         & \multicolumn{2}{c}{98.6\%}   \\\hline
Ours (SVMP)                  & \multicolumn{2}{c}{\textbf{99.3\%}}   \\\hline
\multicolumn{3}{c}{YUP++}                                   \\\hline
Method                       & Stationary     & Moving      \\\hline
Temporal Residual Networks\cite{feichtenhofer2017temporal}&92.4\%         & 81.5\%      \\\hline
Ours (SVMP)&\textbf{92.9\%}&   \textbf{84.0\%}          \\\hline
\end{tabular}}
\end{table}

\subsection{Comparisons to the State of the Art}
In Table \ref{table:10}, we compare our best results against the state-of-the-art on each dataset using the standard evaluation protocols. For a fair comparison, we also report on SVMP combined with hand-crafted features (IDT-FV)~\cite{wang2013dense} for HMDB-51. Our scheme outperforms other methods on all datasets by 1--4\%. For example, on HMDB-51, our results are about 2-3\% better than the next best method without IDT-FV. On Charades, we outperform previous methods by about 3\% while faring well on the detection task against~\cite{Sigurdsson_2017_CVPR}. We also demonstrate significant performance (about 3-4\%) improvement on NTU-RGBD and marginally better performance on MSR datasets on skeleton-based action recognition. Our results are superior (by 1-2\%) on the PubFig and YUP++ datasets. 

We further analyze the benefits of combining I3D+ with SVMP (instead of their proposed average pooling) on both HMDB-51 and UCF-101 datasets using the settings in~\cite{carreira2017quo}. However, we find that the improvement over average pooling in I3D+ is not significant; which we believe is because learning the SVMP descriptor needs to solve a learning problem implicitly, requiring sufficient number of training samples, i.e., number of frames in the sequence. The I3D network uses 64-frame chunks as one sample, thereby reducing the number of samples for SVMP, leading to sub-optimal learning. We analyze this hypothesis in Table~\ref{tab:11}; each column in this table represents performances on a data subset, filtered as per the minimum number of frames in their sequences. As is clear from the table, while SVMP performs on par with I3D+ when the sequences are shorter, it demonstrates significant benefits on subsets having longer sequences.

\section{Conclusion}
\label{sec:conclude}
In this paper, we presented a simple, efficient, and powerful pooling scheme -- SVM pooling -- for video representation learning. We cast the pooling problem in a multiple instance learning framework, and seek to learn useful decision boundaries on video features against background/noise features. We provide an efficient scheme that jointly learns these decision boundaries and the action classifiers on them. Extensive experiments were showcased on eight challenging benchmark datasets, demonstrating state-of-the-art performance. Given the challenging nature of these datasets, we believe the benefits afforded by our scheme is a significant step towards the advancement of recognition systems designed to represent sets of images or videos.

%.the effectiveness of our scheme by outperforming the state of the art on HMDB, while being on par on UCF101. Going forward, an interesting direction will be to investigate alternative formulations of our objective.
\comment{
It summarize actions in video sequences by filtering useful features from a bag of per-frame CNN features. And We propose to use the output of SVM pooling as a descriptor for representing the sequence, namely the SVM Pooled (SVMP) descriptor. Considering SVMP descriptor as a better representation than CNN features, we extended SVMP to NSVMP to deal with the  non-linear sequences. Extensive experiments were implemented on two popular benchmark datasets: HMDB51 and UCF101, which clearly show the advantage and effectiveness of SVMP compared with the traditional pooling scheme on CNN features. More importantly, after combining with the hand-crafted local features, we reach the state-of-the-art result on these two datasets.
}

% \ifCLASSOPTIONcompsoc
%   % The Computer Society usually uses the plural form
%   \section*{Acknowledgments}
% \else
%   % regular IEEE prefers the singular form
% \fi


\ifCLASSOPTIONcaptionsoff
  \newpage
\fi

{\small
\bibliographystyle{IEEEtran}
\bibliography{dbcnn_cvpr}
}

% \end{IEEEbiography}
% \vspace{-1.0cm}
\begin{IEEEbiography}[{\includegraphics[width=1in,height=1.25in,clip,keepaspectratio]{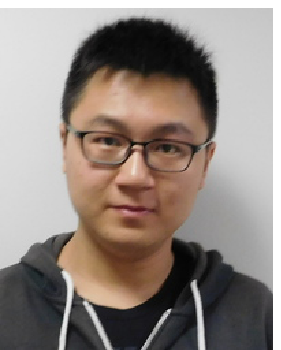}}]{Jue Wang}
is a PhD student with the Research School of Engineering at the Australian National University since 2016. He is also associated with CSIRO's Data61 in Australia. From 2010-2014, he received his double bachelor degree (honors) in Electronic Engineering from Australian National University and Beijing Institute of Technology. His research interest are in the area of computer vision and machine learning.
\end{IEEEbiography}
% \vspace{-1.5cm}
\begin{IEEEbiography}[{\includegraphics[width=1in,height=1.25in,clip,keepaspectratio]{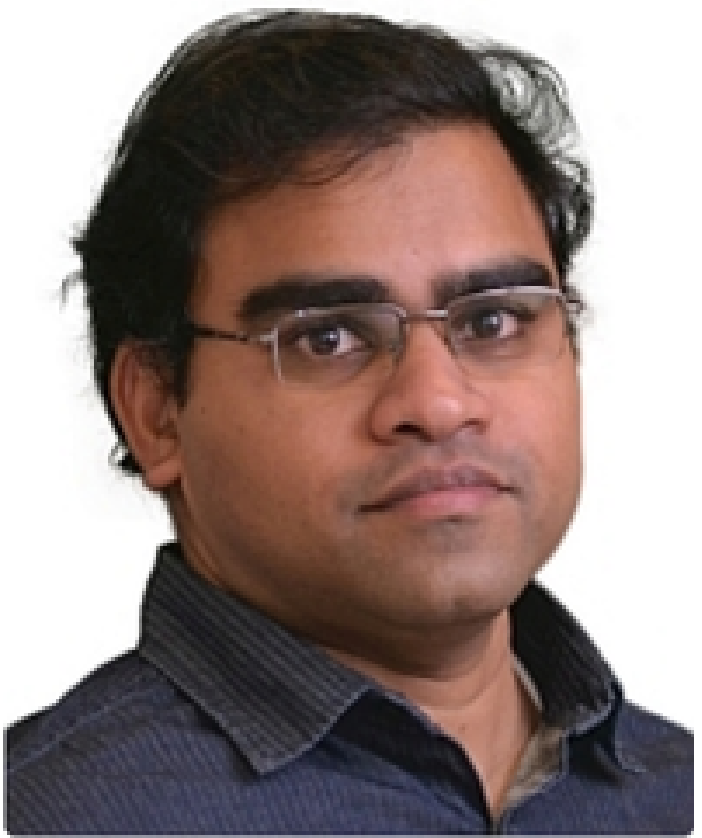}}]{Anoop Cherian}
is a Research Scientist with Mitsubishi Electric Research Labs (MERL) Cambridge, MA and an Adjunct Researcher affiliated to the Australian Centre for Robotic Vision (ACRV) at the Australian National University. Previously, he was a Postdoctoral Researcher in the LEAR team at INRIA at Grenoble. He received his B.Tech (honors) degree in computer science and Engineering from the National Institute of Technology, Calicut, India in 2002, his M.S. and Ph.D. degrees in computer science from the University of Minnesota, Minneapolis in 2010 and 2013 respectively. His research interests lie in the areas of computer vision and machine learning.
\end{IEEEbiography}

%\newpage



%\vfill


\end{document}